\documentclass{article}

 \usepackage[main, final]{neurips_2026}

\usepackage[utf8]{inputenc} 
\usepackage[T1]{fontenc}    
\usepackage{hyperref}       
\usepackage{url}            
\usepackage{booktabs}       
\usepackage{amsfonts}       
\usepackage{nicefrac}       
\usepackage{microtype}      
\usepackage{xcolor}         
\usepackage{graphicx}
\usepackage{subcaption}
\usepackage{hyperref}
\usepackage{amsmath}
\usepackage{amssymb}
\usepackage{mathtools}
\usepackage{amsthm}
\usepackage{multirow}
\usepackage{wrapfig}
\usepackage{bm}
\usepackage{algorithm}
\usepackage{algorithmic}

\newcommand{\sO}{\mathcal{O}}
\newcommand{\sS}{\mathcal{S}}
\newcommand{\sA}{\mathcal{A}}

\newcommand{\INPUT}{\item[\textbf{Input:}]}
\newcommand{\OUTPUT}{\item[\textbf{Output:}]}

\theoremstyle{definition}
\newtheorem{theorem}{Theorem}[section]
\newtheorem{proposition}[theorem]{Proposition}

\newtheorem{corollary}[theorem]{Corollary}
\newtheorem{definition}[theorem]{Definition}
\newtheorem{assumption}[theorem]{Assumption}
\newtheorem{remark}[theorem]{Remark}

\title{Intra-Option Fitted Q-Evaluation: Evaluating Hierarchical Policies from Non-Hierarchical Data}

\author{%
  Yunfu Deng \\
  Department of Computer Sciences\\
  University of Wisconsin--Madison\\
  \texttt{yunfu.deng@wisc.edu}
  \And
  Josiah P.~Hanna \\
  Department of Computer Sciences\\
  University of Wisconsin--Madison\\
  \texttt{jphanna@cs.wisc.edu} 
}

\begin{document}

\maketitle

\begin{abstract}
Off-policy evaluation (OPE) estimates the expected return of a target policy from previously collected data without additional environment interaction. While OPE methods for flat Markovian policies are well studied, little work has addressed evaluating hierarchical policies in which a high-level policy selects temporally extended options that generate primitive actions until a termination condition is met. The primary existing approach applies per-decision importance sampling at the option level, but requires option-annotated trajectories and exhibits variance that still grows with the horizon. For non-hierarchical policies, fitted Q-evaluation (FQE) typically achieves lower error than importance sampling by leveraging the classic Bellman equation for policy evaluation; however, a naive extension of FQE to hierarchical policies results in a biased policy value estimate. In this paper, we introduce Intra-Option Fitted Q-Evaluation (IO-FQE), which estimates a value function in the augmented state-option space; this approach enables OPE of hierarchical policies even when we lack annotations of what option was run in the data. We develop two continuous-action instantiations of IO-FQE and show on long-horizon hierarchical continuous control tasks (AntMaze navigation and OGBench Puzzle manipulation) that IO-FQE eliminates the bias incurred by a naive application of FQE to hierarchical policies and substantially lowers mean squared-error compared to option-level importance sampling.
\end{abstract}

\section{Introduction}
Off-policy evaluation (OPE) estimates the expected return of a target policy from previously collected data without additional environment interaction~\citep{precup2000eligibility, jiang2016doubly, thomas2016data}. Many target policies of practical interest are hierarchical: a high-level policy selects among temporally extended options~\citep{sutton1999between} and each selected option generates primitive actions through an associated sub-policy until a termination condition is met. This structure appears in locomotion with learned skill libraries~\citep{peng2022ase}, manipulation with task-and-motion primitives~\citep{shi2025hi}, and multi-stage navigation with waypoint-following controllers~\citep{park2023hiql}. Despite the prevalence of hierarchical policies, OPE methods have focused almost exclusively on evaluating flat Markovian policies, and the problem of evaluating a hierarchical target policy remains largely unaddressed.

To the best of our knowledge, the only existing approach to hierarchical policy evaluation is that of \citet{guo2017using}, who apply per-decision importance sampling at both high- and low-levels of control. Their estimator is unbiased and consistent, but requires the offline data to have been produced by a hierarchical behavior policy and uses importance weights whose variance grows exponentially with the horizon. When performing OPE on non-hierarchical policies, fitted Q-evaluation (FQE)~\citep{le2019batch} often achieves lower error than importance sampling by replacing trajectory reweighting with iterative  regression to one-step Bellman targets. However, FQE requires a target policy that maps each state to a distribution over primitive actions, and a hierarchical policy whose action distribution also depends on the active option cannot be specified in this form~\citep{sutton1998intra}.

In this paper, we introduce and investigate Intra-Option Fitted Q-Evaluation (IO-FQE), which extends FQE to evaluate hierarchical target policies.
The key idea is to perform Bellman regression in the augmented state-option space $\sS \times \sO$. In this space, the primitive-action distribution under the hierarchical policy depends only on the current state-option pair, so the target policy for FQE is Markov. Consequently, the option-value function $Q^\Pi(s,o)$ satisfies an intra-option Bellman equation~\citep{sutton1998intra}. The option variable is introduced by the estimator rather than read from the dataset: each primitive transition $(s,a,r,s')$ yields a Bellman target for every option in $\sO$, so no option-level annotations are required and we can perform OPE with data generated by non-hierarchical policies. In continuous action spaces, the augmented-space Bellman equation admits two estimation strategies that differ in where the intra-option policy $\pi_o$ enters the computation. IO-FQE-BE learns option-conditioned values $Q(s,o)$ and corrects the mismatch between $\pi_o$ and the data action distribution via importance weighting, estimating the behavior action density from data following regression importance sampling~\citep{hanna2019importance}. IO-FQE-AC learns action-conditioned option values $Q(s,o,a)$ and replaces importance weighting with Monte Carlo marginalization over $\pi_o$, avoiding behavior-density estimation at the cost of querying the learned value function at actions sampled from $\pi_o$ that may lie outside the data support.

This paper makes three central contributions. First, we formalize the irreducible bias incurred by applying flat FQE to a per-step policy that discards option commitment; this bias is a property of the MDP and the hierarchical policy and cannot be reduced with more data or a more expressive function class (Proposition~\ref{prop:irreducible_bias} and Corollary~\ref{cor:irreducibility}). Second, we derive IO-FQE as a Bellman-regression estimator that evaluates the hierarchical policy in the augmented state-option space at the primitive time scale, using only primitive transition data without option-level annotations. 
%
Third, we develop two continuous-action instantiations, IO-FQE-BE and IO-FQE-AC, and evaluate them on AntMaze and OGBench Puzzle; the more accurate variant attains substantially lower MSE than Option-PDIS despite using no option annotations in the behavior data, and flat FQE returns biased estimates due to not accounting for option commitment.
\section{Problem Formulation}
\label{sec:problem_formulation}

In this section, we formalize off-policy evaluation (OPE) of a hierarchical target policy from offline data containing only primitive, one-step transitions.

\subsection{The Primitive MDP and Offline Dataset}

We consider an infinite-horizon discounted MDP $\mathcal{M} = (\sS, \sA, P, R, \gamma, \rho_0)$ with continuous state space $\sS \subseteq \mathbb{R}^d$, continuous primitive action space $\sA \subseteq \mathbb{R}^m$, transition model $P: \sS \times \sA \to \Delta(\sS)$, reward function $R: \sS \times \sA \to \mathbb{R}$, discount factor $\gamma \in [0, 1)$, and initial-state distribution $\rho_0 \in \Delta(\sS)$. In the offline setting, the agent cannot interact with the environment and is instead provided with a static dataset $\mathcal{D} = \{(s_i, a_i, r_i, s'_i)\}_{i=1}^N$ collected by an unknown behavior process, which may consist of one or multiple behavior policies. The behavior action density is not assumed to be known a priori; whether it must be estimated from $\mathcal{D}$ depends on the evaluation strategy adopted in Section~\ref{sec:method}.

\subsection{Offline Evaluation of Hierarchical Target Policies}

Our objective is to use a given offline dataset to estimate the expected cumulative reward that a hierarchical target policy $\Pi$ will obtain when deployed in an MDP environment.
We formalize hierarchical policies using the options framework~\citep{sutton1999between}.

\begin{definition}[Options]
\label{def:options}
Following \citet{sutton1999between}, an option $o$ is a tuple $\langle \mathcal{I}_o, \pi_o, \beta_o \rangle$, where $\mathcal{I}_o \subseteq \sS$ is an initiation set specifying in which states the option can be invoked, $\pi_o(a \mid s)$ is an intra-option policy, and $\beta_o(s) \in [0,1]$ is a termination function giving the probability that the option terminates upon arrival in state $s$.
\end{definition}

\paragraph{Time-dependent termination}
The termination function $\beta_o$ defined above depends only on the arrival state. A termination condition that also depends on the time elapsed since the option was initiated is handled by augmenting the state with the elapsed time $h$. On the augmented state $(s, o, h)$, termination is Markov: continuation increments $h$, whereas termination selects a new option and resets $h$ to zero. 
As with the option identity, IO-FQE enumerates $h$ internally rather than reading it from the data, so Assumption~\ref{assump:flat_data} remains unchanged. For options of bounded duration, this augmentation adds one scalar input under function approximation and enlarges the tabular state space by the modeled duration range. Appendix~\ref{app:time_dependent_termination} evaluates this extension empirically.

Let $\sO = \{o_1, \dots, o_K\}$ denote the set of available options. A hierarchical policy is specified by a high-level policy $\mu(o \mid s)$ over $\sO$, in which $\mu(\cdot \mid s)$ assigns probability mass only to feasible options ($\mu(o \mid s) = 0$ if $s \notin \mathcal{I}_o$). 
The resulting hierarchical target policy is denoted $\Pi \triangleq (\mu, \sO)$. Executing $\Pi$ in $\mathcal{M}$ following the call-and-return procedure of \citet{sutton1999between} generates a random sequence $(s_t, o_t, a_t, r_t)_{t \geq 0}$ with $s_0 \sim \rho_0$, $o_0 \sim \mu(\cdot \mid s_0)$, and for each $t \geq 0$, $a_t \sim \pi_{o_t}(\cdot \mid s_t)$, $r_t = R(s_t, a_t)$, $s_{t+1} \sim P(\cdot \mid s_t, a_t)$, and
\begin{equation}
o_{t+1} \sim 
\begin{cases}
\mu(\cdot \mid s_{t+1}) & \text{with probability } \beta_{o_t}(s_{t+1}), \\
\delta_{o_t} & \text{with probability } 1 - \beta_{o_t}(s_{t+1}).
\end{cases}
\end{equation}
The expected discounted return of $\Pi$ is $J(\Pi) \triangleq \mathbb{E}_{s_0 \sim \rho_0,\, \Pi}\!\left[\sum_{t=0}^{\infty} \gamma^t r_t\right]$, where the expectation is over the trajectory distribution induced by the above procedure. The goal of OPE is to estimate $J(\Pi)$ using only the offline dataset $\mathcal{D}$.

\begin{assumption}[Flat data constraint]
\label{assump:flat_data}
The dataset $\mathcal{D}$ contains only transitions $(s,a,r,s')$ and provides no option-level annotations (e.g., option identities or initiation/termination events) or any latent state of the data-collection process.
\end{assumption}

\begin{figure}[t]
\centering
\includegraphics[width=\linewidth]{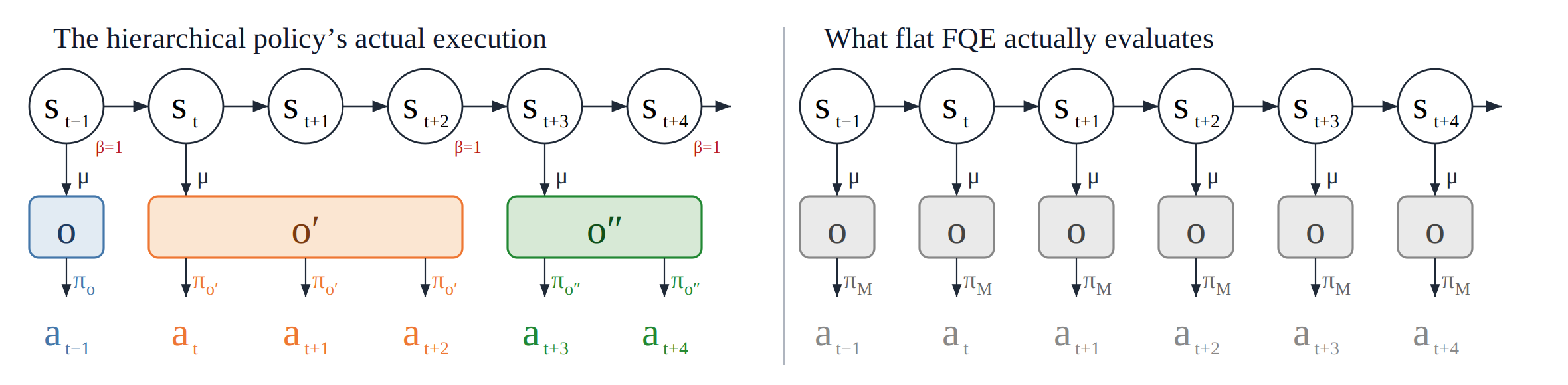}
\caption{The committed hierarchical policy $\Pi$ (left) executes an option until termination is sampled from $\beta_o(\cdot)$ before drawing a new one from $\mu(\cdot \mid s)$, whereas $\pi_M$ (right), the policy flat FQE actually evaluates, resamples an option from $\mu(\cdot \mid s)$ at every primitive step. The two induce different actions along the same state trajectory; the resulting return gap $\Delta(\Pi) = |J(\Pi) - J(\pi_M)|$ is formalized in Proposition~\ref{prop:irreducible_bias}.}
\label{fig:wrong_target}
\end{figure}

\paragraph{Hierarchical policies are Non-Markov} 
Executing $\Pi$ in the primitive MDP produces a process in which the primitive action distribution at each step depends on the active option, which in turn depends on the history of option initiation and termination events~\citep{sutton1999between}. Standard fitted Q-evaluation~\citep{le2019batch} takes as input a target policy that maps each state to a distribution over primitive actions; because the primitive-level action distribution under $\Pi$ also depends on the active option, $\Pi$ cannot be expressed as a mapping from states alone to action distributions. 
A natural alternative is the policy
\begin{equation}
\label{eq:pi_M}
\pi_M(a \mid s) \triangleq \sum_{o \in \sO} \mu(o \mid s)\,\pi_o(a \mid s),
\end{equation}
which selects an option from $\mu(\cdot \mid s)$ at every primitive step rather than maintaining option commitment across steps. While $\pi_M$ is Markov by construction, $J(\pi_M) \neq J(\Pi)$ in general, because the multi-step commitment that $\Pi$ enforces within each option is absent from $\pi_M$ (Figure~\ref{fig:wrong_target}).

Given a flat Markov target policy $\pi$, FQE learns an action-value approximation $\hat{Q}(s, a)$ through iterative Bellman regression and estimates the policy value as
\begin{equation}
    \label{eq:fqe_estimate}
    \hat{J}_{\mathrm{FQE}}(\pi;\, \mathcal{D}) \triangleq \frac{1}{|\mathcal{D}_0|} \sum_{s_0 \in \mathcal{D}_0} \hat{V}_\pi(s_0), \quad \text{where } \hat{V}_\pi(s) = \mathbb{E}_{a \sim \pi(\cdot \mid s)}\!\left[\hat{Q}(s, a)\right]
\end{equation}
and $\mathcal{D}_0$ denotes the set of initial states in $\mathcal{D}$. When applied to $\pi_M$, the estimator is a consistent estimator of $J(\pi_M)$ rather than $J(\Pi)$.

\begin{proposition}[Option commitment gap of flat FQE]
\label{prop:irreducible_bias}
Let $\Pi = (\mu, \sO)$ be a hierarchical policy, let $\pi_M$ be the Markov policy defined in Eq.~\eqref{eq:pi_M}, and let $\Delta(\Pi) \triangleq |J(\pi_M) - J(\Pi)|$ denote the option commitment gap. For any dataset $\mathcal{D}$ satisfying Assumption~\ref{assump:flat_data} and any FQE estimator $\hat{J}_{\mathrm{FQE}}(\cdot; \mathcal{D})$ that takes a Markov target policy as input, let $\epsilon_{\mathrm{est}}(\mathcal{D}) \triangleq \big|\hat{J}_{\mathrm{FQE}}(\pi_M; \mathcal{D}) - J(\pi_M)\big|$ denote the estimation error of the estimator with respect to its own target $J(\pi_M)$. The error in evaluating $J(\Pi)$ then satisfies
\begin{equation}
\label{eq:bias_decomp}
\Delta(\Pi) - \epsilon_{\mathrm{est}}(\mathcal{D}) \;\le\; \big|\hat{J}_{\mathrm{FQE}}(\pi_M; \mathcal{D}) - J(\Pi)\big| \;\le\; \Delta(\Pi) + \epsilon_{\mathrm{est}}(\mathcal{D}).
\end{equation}
\end{proposition}

\begin{corollary}[Irreducibility of the option commitment gap]
\label{cor:irreducibility}
Suppose the FQE estimator is consistent for estimating $J(\pi_M)$: for datasets $\mathcal{D}_N$ of size $N$, $\epsilon_{\mathrm{est}}(\mathcal{D}_N) \to 0$ with probability $1$ as $N \to \infty$. Then $\big|\hat{J}_{\mathrm{FQE}}(\pi_M; \mathcal{D}_N) - J(\Pi)\big| \to \Delta(\Pi)$ with probability $1$.
\end{corollary}

A formal proof of both statements is given in Appendix~\ref{app:proofs_sec2}. The estimation error $\epsilon_{\mathrm{est}}(\mathcal{D})$ depends on the dataset, the function class, and the optimization procedure. The option commitment gap $\Delta(\Pi)$, by contrast, is a property of the pair $(\mathcal{M}, \Pi)$ alone and does not depend on $\mathcal{D}$, the function class, or the optimization procedure. Eq.~\eqref{eq:bias_decomp} therefore confines the error of flat FQE to an interval of width $2\epsilon_{\mathrm{est}}(\mathcal{D})$ centered at $\Delta(\Pi)$: no increase in dataset size or approximation capacity reduces the error below $\Delta(\Pi) - \epsilon_{\mathrm{est}}(\mathcal{D})$, and under the consistency premise of Corollary~\ref{cor:irreducibility}, the standard premise of FQE analyses under suitable realizability, Bellman-completeness, coverage, and estimation conditions~\citep{le2019batch, munos2008finite}, the error converges to $\Delta(\Pi)$ itself. Appendix~\ref{app:ablation_tabular} validates the corollary empirically: in a tabular setting in which $\epsilon_{\mathrm{est}}$ is driven to numerical precision, the flat-FQE error plateaus at $\Delta(\Pi)$ across all dataset sizes.

To remove the target mismatch between $\pi_M$ and $\Pi$, the evaluation must maintain option identity throughout, operating in the augmented space $\sS \times \sO$ rather than in $\sS$ alone. Section~\ref{sec:method} develops this approach based on intra-option learning \citep{sutton1998intra}.
\section{Intra-Option Fitted Q-Evaluation}
\label{sec:method}
Corollary~\ref{cor:irreducibility} establishes that flat FQE incurs an option commitment gap $\Delta(\Pi)$ when targeting $J(\Pi)$. IO-FQE eliminates this bias by lifting evaluation to the augmented space $\sS \times \sO$, in which the process $\{(s_t, o_t)\}$ under $\Pi$ is Markov: given $(s_t, o_t)$ and the primitive action $a_t$, the next state follows $P(\cdot \mid s_t, a_t)$ and the option either continues with probability $1 - \beta_{o_t}(s_{t+1})$ or terminates and triggers reselection from $\mu(\cdot \mid s_{t+1})$. The same holds on $(s, o, h)$ under time-dependent termination.

In this section, we introduce IO-FQE, which operates at the primitive time scale using the intra-option value functions of \citet{sutton1998intra}. Let $Q^\Pi(s, o)$ denote the expected return when option $o$ is initiated in state $s$ and the agent follows $\Pi$ thereafter. The high-level value and value upon arrival are
\begin{align}
\label{eq:high_level_value}
V^\Pi(s) &\triangleq \textstyle\sum_{o'} \mu(o' \mid s)\, Q^\Pi(s, o'), \\
\label{eq:arrival_value}
U^\Pi(s, o) &\triangleq (1 - \beta_o(s))\, Q^\Pi(s, o) + \beta_o(s)\, V^\Pi(s),
\end{align}
where $U^\Pi$ reflects whether option $o$ continues or terminates. The option-value function satisfies the intra-option Bellman equation:
\begin{equation}
\label{eq:bellman_so}
Q^\Pi(s, o) = \mathbb{E}_{a \sim \pi_o(\cdot \mid s)}\!\Big[R(s,a) + \gamma\,\mathbb{E}_{s' \sim P(\cdot \mid s,a)}\!\big[U^\Pi(s', o)\big]\Big],
\end{equation}
in which the expectation is taken under the intra-option policy $\pi_o$. Each transition $(s,a,r,s') \in \mathcal{D}$ yields a Bellman target for every option simultaneously, since the right-hand side involves only primitive-level quantities and requires no option annotation. Section~\ref{subsec:targets} constructs the temporal-difference targets. How $Q_\theta(s,o)$ is learned from primitive data admits two strategies: IO-FQE-BE (Section~\ref{subsec:iofqe_be}) corrects the mismatch between $\pi_o$ and the data action density via importance weighting; IO-FQE-AC (Section~\ref{subsec:iofqe_ac}) avoids it by conditioning on the primitive action. Appendix~\ref{app:theory} provides Bellman consistency proofs and finite-sample bounds for both.

\subsection{Bellman Targets from Primitive Transitions}
\label{subsec:targets}

Let $Q_\theta(s, o)$ denote a parametric approximation of the option-value function $Q^\Pi(s, o)$. Substituting $Q_\theta$ for $Q^\Pi$ in Eqs.~\eqref{eq:high_level_value} and~\eqref{eq:arrival_value} yields the high-level value $V_\theta(s) \triangleq \sum_{o \in \sO} \mu(o \mid s)\, Q_\theta(s, o)$ and the termination-aware arrival value $U_\theta(s, o) \triangleq (1 - \beta_o(s))\, Q_\theta(s, o) + \beta_o(s)\, V_\theta(s)$. For each transition $(s, a, r, s') \in \mathcal{D}$ and each option $o \in \sO$, the temporal-difference target is
\begin{equation}
    \label{eq:td_target}
    y_o \triangleq r + \gamma \, U_{\bar{\theta}}(s', o),
\end{equation}
where $\bar{\theta}$ denotes the parameters of a periodically updated target function \citep{mnih2015human}.

The construction above defines the target $y_o$ in terms of any approximation $Q_\theta(s, o)$, but does not specify how $Q_\theta(s, o)$ is itself learned from primitive transitions. The option-conditioned Bellman form (Eq.~\eqref{eq:bellman_so}) provides a natural starting point: it characterizes $Q^\Pi(s, o)$ directly, suggesting a regression of $Q_\theta(s, o)$ against the targets $y_o$. 

The central difficulty in this regression is that the Bellman equation (Eq.~\eqref{eq:bellman_so}) takes an expectation over actions drawn from the intra-option policy $\pi_o$, whereas the actions recorded in $\mathcal{D}$ are drawn from the behavior action density $p_\mathcal{D}(a \mid s)$. Unless this distribution mismatch is corrected, the fixed point of the regression does not recover $Q^\Pi(s, o)$. The following two subsections develop complementary strategies for resolving this mismatch: IO-FQE-BE (Section~\ref{subsec:iofqe_be}) corrects for it via importance weighting, requiring estimation of $p_\mathcal{D}$ from data; IO-FQE-AC (Section~\ref{subsec:iofqe_ac}) avoids it by conditioning the value function on the primitive action, eliminating importance weights at the cost of Monte Carlo marginalization over $\pi_o$.

\subsection{IO-FQE via Behavior Policy Estimation}
\label{subsec:iofqe_be}

The first variant, IO-FQE with Behavior Estimation (IO-FQE-BE), learns $Q_\theta(s, o)$ by regressing against the targets $y_o$ (Eq.~\eqref{eq:td_target}). In standard FQE, the Q-function $Q(s, a)$ conditions on the primitive action, and the target policy enters only through the next-state value $V(s') = \sum_{a'} \pi(a' \mid s')\, Q(s', a')$; no importance weighting of the loss is needed. IO-FQE-BE conditions on the option rather than the action, so the mismatch between $\pi_o$ and the data action distribution must be corrected explicitly. Without correction, the regression takes the Bellman expectation under $p_\mathcal{D}(a \mid s)$ rather than under $\pi_o(a \mid s)$, and the resulting fixed point does not in general equal $Q^\Pi(s, o)$. To correct this mismatch, IO-FQE-BE introduces the importance ratio
\begin{equation}
    \label{eq:importance_weight}
    w_o(s, a) \triangleq \frac{\pi_o(a \mid s)}{p_\mathcal{D}(a \mid s)},
\end{equation}
which performs the change of measure from $p_\mathcal{D}(\cdot \mid s)$ to $\pi_o(\cdot \mid s)$. The IO-FQE-BE objective is the importance-weighted mean-squared Bellman error:
\begin{equation}
    \label{eq:io_fqe_be_loss}
    \mathcal{L}_{\mathrm{BE}}(\theta) = \mathbb{E}_{(s,a,r,s') \sim \mathcal{D}} \left[ \sum_{o \in \sO} w_o(s, a) \left( Q_\theta(s, o) - y_o \right)^2 \right].
\end{equation}
Eq.~\eqref{eq:io_fqe_be_loss} is a weighted least-squares problem whose population minimizer satisfies the intra-option Bellman equation. Fix $(s, o)$ and take the expectation of the weighted squared error given $s$, over $a \sim p_\mathcal{D}(\cdot \mid s)$ and $s' \sim P(\cdot \mid s, a)$. Because $Q_\theta(s, o)$ does not depend on $a$, the minimizer of this expectation over an unrestricted function class is the $w_o$-weighted conditional mean of the target, $\mathbb{E}_{a \sim p_\mathcal{D}}[w_o(s, a)\, y_o \mid s]$, which needs no normalization because $\mathbb{E}_{a \sim p_\mathcal{D}}[w_o(s, a) \mid s] = 1$; by the change of measure in Eq.~\eqref{eq:importance_weight}, this mean equals $\mathbb{E}_{a \sim \pi_o}[y_o \mid s]$, the right-hand side of Eq.~\eqref{eq:bellman_so} with $Q_{\bar\theta}$ as the bootstrap function. Proposition~\ref{prop:bellman_consistency_be} in Appendix~\ref{app:theory} states this result, and Remark~\ref{rem:weight_placement} compares weighting the squared error with the alternative of applying $w_o$ to the target $y_o$, which has the same population minimizer.

Following regression importance sampling~\citep{hanna2019importance, pmlr-v119-pavse20a}, IO-FQE-BE estimates $p_\mathcal{D}$ from $\mathcal{D}$ itself, which allows the importance weights to correct for sampling error in the observed action distribution.
The density model can be parameterized by any sufficiently expressive function class (e.g., a Gaussian policy network or a mixture density network). For the importance weights to be well-defined, $\pi_o(\cdot \mid s)$ must be absolutely continuous with respect to $\hat{p}_\mathcal{D}(\cdot \mid s)$ for all $s$ in the support of the data distribution and all $o \in \sO$; this is the standard coverage requirement in off-policy evaluation~\citep{precup2000eligibility}. The model is estimated once before the main evaluation loop and held fixed throughout training. 

In high-dimensional continuous action spaces, however, accurate density estimation becomes challenging as the expressiveness required of the model grows with the action dimensionality, and misspecification can degrade the importance weights. 
With the true behavior action density, the Bellman operator defined by the minimizer of $\mathcal{L}_{\mathrm{BE}}$ is a $\gamma$-contraction with unique fixed point $Q^\Pi$ (Proposition~\ref{prop:bellman_consistency_be}). With an estimated density $\hat{p}_\mathcal{D}$, the weight perturbation introduces an additive bias of order $\epsilon_p \cdot R_{\max} / (1 - \gamma)^2$ that does not vanish with increasing dataset size (Theorem~\ref{thm:error_be}). To avoid importance weighting and the associated density estimation requirement entirely, Section~\ref{subsec:iofqe_ac} develops an alternative, IO-FQE-AC (Action-Conditioned), based on the action-conditioned Bellman form (Eq.~\eqref{eq:bellman_soa}).

\subsection{IO-FQE via Action Conditioning}
\label{subsec:iofqe_ac}

The second variant, IO-FQE with Action-Conditioned Regression (IO-FQE-AC), avoids importance weighting by learning the action-conditioned option-value function $Q_\theta(s, o, a)$ directly. The value $Q^\Pi(s, o, a)$ satisfies the action-conditioned intra-option Bellman equation:
\begin{equation}
\label{eq:bellman_soa}
Q^\Pi(s, o, a) = R(s, a) + \gamma \, \mathbb{E}_{s' \sim P(\cdot \mid s, a)} \!\big[ U^\Pi(s', o) \big],
\end{equation}
in which no expectation over actions appears on the right-hand side. Let $Q_\theta: \sS \times \sA \to \mathbb{R}^{|\sO|}$ be a parametric function approximator that takes a state-action pair as input and outputs action-conditioned option-values for all options simultaneously. Because $Q_\theta$ conditions on the primitive action recorded in each transition, no mismatch between $\pi_o$ and the data action distribution arises in the regression, and the training objective is the unweighted mean-squared Bellman error:
\begin{equation}
    \label{eq:io_fqe_ac_loss}
    \mathcal{L}_{\mathrm{AC}}(\theta) = \mathbb{E}_{(s,a,r,s') \sim \mathcal{D}} \left[ \sum_{o \in \sO} \left( Q_\theta(s, o, a) - y_o \right)^2 \right].
\end{equation}
No importance weights or behavior density estimates appear in this objective.

The target policy $\pi_o$ enters IO-FQE-AC through the target computation rather than through the loss. Computing the target $y_o = r + \gamma\, U_{\bar\theta}(s', o)$ requires the option-value $Q_{\bar\theta}(s', o)$, which is recovered from the action-conditioned value by marginalizing over the intra-option policy: $Q_\theta(s, o) = \mathbb{E}_{a \sim \pi_o(\cdot \mid s)}[Q_\theta(s, o, a)]$. In continuous action spaces, this expectation is approximated by drawing $M$ samples $\{a'_m\}_{m=1}^M \sim \pi_o(\cdot \mid s')$ and computing $\hat{Q}_{\bar\theta}(s', o) = \frac{1}{M} \sum_{m=1}^{M} Q_{\bar\theta}(s', o, a'_m)$. Under bounded rewards, $|Q^\Pi(s, o, a)| \le R_{\max}/(1-\gamma)$, so the variance of each per-target Monte Carlo estimate $\hat{Q}_{\bar\theta}(s', o)$ is at most $R_{\max}^2 / ((1-\gamma)^2 M)$ and decreases as $O(1/M)$. 
Proposition~\ref{prop:bellman_consistency_ac} in Appendix~\ref{app:theory} establishes that the population minimizer of $\mathcal{L}_{\mathrm{AC}}$ recovers the action-conditioned intra-option Bellman equation (Eq.~\eqref{eq:bellman_soa}) and the resulting operator is a $\gamma$-contraction with unique fixed point $Q^\Pi$.
The MC marginalization error enters the finite-sample bound (Theorem~\ref{thm:error_ac}) as an additive term of order $R_{\max} / ((1 - \gamma)^2 \sqrt{M})$ in expectation, controlled entirely by the computational budget $M$ and independent of the data distribution.

\paragraph{Summary of the tradeoff.}
Both IO-FQE-BE and IO-FQE-AC converge to $Q^\Pi$ when the regression is solved exactly over an unrestricted function class, provided the behavior data covers the support of every intra-option policy $\pi_o$ (Assumptions~\ref{assump:coverage_be} and~\ref{assump:coverage_ac}; IO-FQE-BE additionally requires bounded density ratios), but their finite-sample error bounds (Theorems~\ref{thm:error_be} and~\ref{thm:error_ac}) expose complementary error sources. IO-FQE-BE avoids querying the learned value function at actions outside the data support, but introduces a density estimation bias $\epsilon_p$ that persists regardless of dataset size. IO-FQE-AC eliminates density estimation entirely, but queries the value function at actions sampled from $\pi_o$ during bootstrapping, incurring an extrapolation risk that grows with the distributional shift between the state-action occupancy of $\Pi$ and the data coverage. The MC marginalization error decreases as $O(1/\sqrt{M})$ and can be reduced by increasing the sample count. Appendix~\ref{app:theory} provides full proofs of Bellman consistency and finite-sample error bounds for both variants, along with a comparison of their error structures with that of flat FQE.
\section{Experiments}
\label{sec:experiments}

The experiments address two questions. 

\begin{enumerate}
    \setlength{\itemsep}{0pt}\setlength{\parskip}{0pt}\setlength{\topsep}{0pt}
    \item When option-annotated trajectories are available, does IO-FQE provide a more accurate estimate of $J(\Pi)$ than Option-PDIS~\citep{guo2017using}?
    \item Does IO-FQE remain accurate when only flat primitive data is available and Option-PDIS is inapplicable?
\end{enumerate}
To answer these questions, we construct two dataset regimes: a hierarchical regime in which data is collected by a hierarchical behavior policy with option commitment (option annotations available to Option-PDIS), and a non-hierarchical flat regime in which the behavior policy is a flat primitive policy (no option annotations available). Flat FQE is included as an additional reference to empirically confirm the option commitment gap $\Delta(\Pi)$ characterized in Corollary~\ref{cor:irreducibility}. Section~\ref{subsec:setup} describes the experimental setup, Section~\ref{subsec:results} presents the main results, and Section~\ref{subsec:data_efficiency} evaluates data efficiency.

\subsection{Experimental Setup}
\label{subsec:setup}

\paragraph{Environments.}
The main evaluation uses two benchmark suites. AntMaze provides sparse-reward long-horizon control with a quadruped agent (observation dimension $d=29$, action dimension $m=8$) navigating from a fixed start state to a goal location. Three maze layouts of increasing size are used: UMaze~\cite{fu2020d4rl} and Medium and Large from OGBench~\cite{park2024ogbench}. OGBench Puzzle provides dense-reward long-horizon control in a combinatorial object-state space. A robot arm uses a five-dimensional relative end-effector action to change the state of a Lights Out puzzle by pressing buttons, in which each press toggles the selected button and its neighbors. The per-step reward equals the negative number of buttons whose states differ from the goal state. Two grid sizes are used: $3\!\times\!3$ and $4\!\times\!6$, with observation dimensions $d=55$ and $d=115$, respectively, and action dimension $m=5$. The combinatorial state space grows up to $2^{24}$ distinct button configurations.

\paragraph{Option and policy construction.}
In AntMaze, the option set consists of four directional locomotion options corresponding to movement toward adjacent waypoints, and the high-level policy $\mu$ is a waypoint-based breadth-first-search policy with $\epsilon$-greedy option selection. In OGBench Puzzle, the option set consists of three manipulation primitives (Move, Press, and Lift) that position, actuate, and retract the end effector, respectively. The high-level policy $\mu$ is a rule-based state machine that sequences these primitives according to a Lights Out solver. For each environment, the target and behavior hierarchical policies share the same option structure but use different intra-option policies, so that $\pi_o^{\text{target}} \neq \pi_o^{\text{behavior}}$ for each option $o$. Full construction details are provided in Appendix~\ref{app:experimental_details}.

\paragraph{Dataset regimes.}
Two dataset regimes are used to separate the option-commitment gap $\Delta(\Pi)$ from the additional estimation effects introduced by hierarchical data collection. In the flat regime, a flat primitive policy is trained via offline RL on existing datasets for each environment, producing data with no option commitment structure. In the hierarchical regime, data are collected by executing a hierarchical behavior policy that maintains option commitment across multiple primitive steps. Both regimes yield datasets of primitive transitions $(s, a, r, s')$ with no option-level annotations, satisfying Assumption~\ref{assump:flat_data}. Dataset sizes and behavior policy details are reported in Appendix~\ref{app:experimental_details}.

\paragraph{Evaluation protocol.}
Ground-truth values $J(\Pi)$ are estimated by Monte Carlo rollouts of $\Pi$ in each environment. All methods are evaluated by mean squared error (MSE) against these values, mean with 95\% bootstrapped confidence intervals over 10 seeds, with additional protocol details in Appendix~\ref{app:experimental_details}.

\paragraph{Methods under comparison.}
All methods receive the same target policy $\Pi = (\mu, \sO)$ and are evaluated against the same ground truth $J(\Pi)$. Option-PDIS~\citep{guo2017using} is equipped with INCRIS for variance reduction in continuous action spaces. Flat FQE is evaluated against both $J(\Pi)$ and $J(\pi_M)$ to separate the option commitment gap from estimation error, with $J(\pi_M)$ estimated by Monte Carlo rollouts. IO-FQE-BE and IO-FQE-AC are as developed in Sections~\ref{subsec:iofqe_be} and~\ref{subsec:iofqe_ac}. All FQE-based estimators share one configuration fixed from standard defaults before any evaluation; the mixture density network of IO-FQE-BE is selected per environment by held-out likelihood alone, and Option-PDIS with INCRIS has no manually tuned hyperparameters (Appendix~\ref{app:hyperparameters}).

\subsection{Results}
\label{subsec:results}

\begin{table}[t]
\centering
\footnotesize
\caption{MSE of policy value estimates ($\downarrow$). Ground-truth values are obtained by Monte Carlo rollout and reported in Appendix~\ref{app:ground_truth}. Under the hierarchical regime, data are collected by a hierarchical behavior policy with option commitment; under the flat regime, by a flat primitive policy. Option-PDIS requires option-annotated trajectories and is inapplicable in the flat regime (marked ---). For Puzzle environments, MSE values are scaled by $10^{-3}$. Values report mean with 95\% bootstrapped confidence intervals over 10 seeds. Paired-bootstrap comparisons between methods are reported in Appendix~\ref{app:significance}.}
\label{tab:results}
\begin{tabular}{@{}ll ccc cc@{}}
\toprule
& & & & & \multicolumn{2}{c}{Flat FQE} \\
\cmidrule(l){6-7}
Regime & Environment & Option-PDIS & IO-FQE-BE & IO-FQE-AC & vs $J(\Pi)$ & vs $J(\pi_M)$ \\
\midrule
\multirow{5}{*}{Hierarchical}
& AntMaze-U       & $6.35 \pm 1.76$   & $0.14 \pm 0.10$  & $0.39 \pm 0.31$ & $1.07 \pm 0.34$   & $0.23 \pm 0.04$ \\
& AntMaze-M       & $55.5 \pm 3.9$    & $0.74 \pm 0.14$  & $0.66 \pm 0.35$ & $7.52 \pm 1.63$   & $1.09 \pm 0.23$ \\
& AntMaze-L       & $79.2 \pm 2.5$    & $1.51 \pm 0.91$  & $0.84 \pm 0.49$ & $14.85 \pm 1.29$  & $1.96 \pm 0.25$ \\
\cmidrule(l){2-7}
& Puzzle-$3\!\times\!3$ {\scriptsize($\times 10^3$)}  & $7.2 \pm 3.8$  & $10.6 \pm 4.4$ & $0.6 \pm 0.3$ & $5.3 \pm 1.2$ & $0.7 \pm 0.1$ \\
& Puzzle-$4\!\times\!6$ {\scriptsize($\times 10^3$)}  & N/A & $436.1 \pm 54.1$ & $46.4 \pm 21.4$ & $301.7 \pm 56.6$ & $67.4 \pm 19.8$ \\
\midrule
\multirow{5}{*}{Flat}
& AntMaze-U       & ---  & $0.05 \pm 0.04$  & $0.40 \pm 0.25$ & $1.05 \pm 0.32$  & $0.74 \pm 0.16$ \\
& AntMaze-M       & ---  & $0.24 \pm 0.06$  & $0.65 \pm 0.35$ & $5.50 \pm 1.47$  & $1.12 \pm 0.16$ \\
& AntMaze-L       & ---  & $0.32 \pm 0.11$  & $0.89 \pm 0.50$ & $13.18 \pm 1.46$ & $1.47 \pm 0.34$ \\
\cmidrule(l){2-7}
& Puzzle-$3\!\times\!3$ {\scriptsize($\times 10^3$)}  & ---  & $0.5 \pm 0.2$ & $0.9 \pm 0.6$ & $3.9 \pm 1.3$ & $1.0 \pm 0.5$ \\
& Puzzle-$4\!\times\!6$ {\scriptsize($\times 10^3$)}  & ---  & $23.8 \pm 15.6$ & $15.5 \pm 9.3$ & $37.4 \pm 12.4$ & $53.1 \pm 16.2$ \\
\bottomrule
\end{tabular}
\end{table}

Table~\ref{tab:results} answers the first question affirmatively: under hierarchical data, the more accurate of the two IO-FQE variants substantially outperforms Option-PDIS on every environment despite Option-PDIS having access to option-annotated trajectories unavailable to IO-FQE. Paired bootstrap comparisons over the shared dataset seeds show that, in each environment and regime, the more accurate IO-FQE variant significantly outperforms every applicable baseline evaluated against $J(\Pi)$, with the 95\% confidence interval of each paired MSE difference excluding zero (Appendix~\ref{app:significance}). Option-PDIS exhibits MSE increasing from $6.35$ on AntMaze-U to $79.2$ on AntMaze-L, and on Puzzle-$4\!\times\!6$, the cumulative importance product over the extended option sequences diverges numerically, producing estimates too unstable to yield meaningful MSE values (marked N/A). These results illustrate the difficulty of trajectory-level importance sampling in long-horizon continuous-action settings: per-step action-density ratios compound across the trajectory, and the resulting variance renders the estimator unreliable even with the Option-PDIS variance reduction of \citet{guo2017using} and privileged access to option-level annotations. In a short-horizon four-rooms gridworld with discrete actions, the confidence intervals of both IO-FQE variants lie below that of Option-PDIS at the $15\!\times\!15$ scale, whereas at the $9\!\times\!9$ scale the interval of Option-PDIS overlaps those of both IO-FQE variants, although its mean MSE remains higher (Appendix~\ref{app:short_horizon}). Option-PDIS is thus a strong choice when the effective horizon is short and option-annotated data are available, while IO-FQE remains competitive using only primitive transitions. Under flat data, the setting of the second question, Option-PDIS is inapplicable entirely, whereas both IO-FQE variants produce accurate estimates across all environments.

The relative performance of the two IO-FQE variants differs across regimes. Under flat data, IO-FQE-BE is more accurate than IO-FQE-AC on every environment except Puzzle-$4\!\times\!6$: the flat behavior policy provides broad action coverage at each state, and a well-specified density model yields bounded importance weights. Under hierarchical data, IO-FQE-BE degrades: on Puzzle-$3\!\times\!3$, its MSE rises from $0.5$ (flat) to $10.6$ (hierarchical), and on Puzzle-$4\!\times\!6$, from $23.8$ to $436.1$, exceeding flat FQE's error against $J(\Pi)$. This degradation reflects a breakdown of the bounded density ratio condition required by IO-FQE-BE. The deterministic high-level state machine produces near-disjoint intra-option action distributions, so $p_\mathcal{D}(a \mid s)$ under hierarchical data assigns negligible density to actions favored by options other than the one that generated the data at state $s$, causing the importance weights to diverge. IO-FQE-AC is unaffected by this condition, achieving MSE of $0.6$ (Puzzle-$3\!\times\!3$) and $46.4$ (Puzzle-$4\!\times\!6$) under the same data, because neither its regression objective nor its Monte Carlo marginalization involves the behavior action density. The practical implication is that when the behavior process is known to be flat or the action space is low-dimensional, IO-FQE-BE offers tighter estimates; when the behavior process is hierarchical or unknown, IO-FQE-AC is the safer default. Appendix~\ref{app:ablation_density} provides an ablation isolating the effect of density estimation quality on IO-FQE-BE.

The two flat FQE columns confirm the option commitment gap characterized in Corollary~\ref{cor:irreducibility}. On AntMaze, flat FQE estimates $J(\pi_M)$ with MSE between $0.23$ and $1.96$, but its error against $J(\Pi)$ ranges from $1.05$ to $14.85$, confirming that flat FQE is an accurate estimator of the wrong target. This pattern holds across both dataset regimes: the dominant error source is the mismatch between $\Pi$ and $\pi_M$, not the structure of the behavior policy.

\subsection{Data Efficiency}
\label{subsec:data_efficiency}

\begin{wrapfigure}{R}{0.50\textwidth}
\vspace{-1em}
\centering
\includegraphics[width=\linewidth]{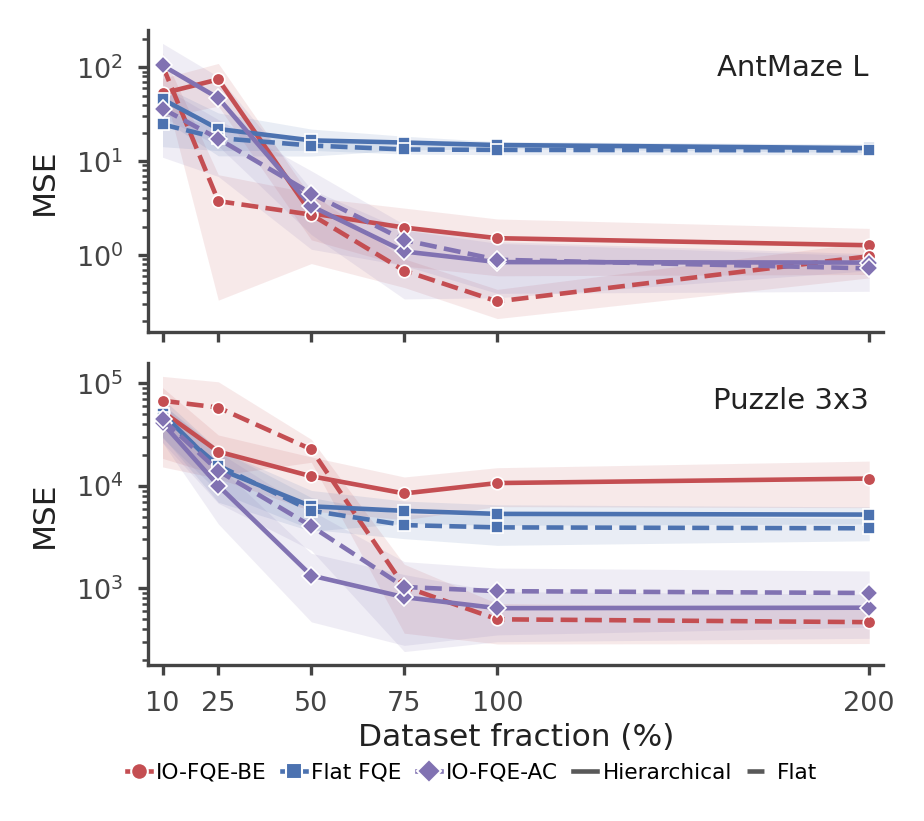}
\caption{MSE vs.\ dataset size on AntMaze-L and Puzzle-$3\!\times\!3$ under flat and hierarchical data regimes. Dataset size varies from 10\% to 200\% of the full dataset. Shaded regions indicate 95\% bootstrapped confidence intervals over 10 seeds.}
\label{fig:data_efficiency}
\vspace{-3.0em}
\end{wrapfigure}

The two data regimes differ in the geometry of action-space coverage: flat behavior distributes data broadly across the action space at each state, whereas hierarchical behavior concentrates data along narrow option-committed trajectories. This subsection examines MSE as a function of dataset size $N$ to characterize how these coverage properties affect scaling behavior.

Figure~\ref{fig:data_efficiency} plots MSE against dataset size for all methods on AntMaze-L and Puzzle-$3\!\times\!3$ under both data regimes. The 200-episode dataset of the main experiments serves as the reference size (100\%); each method is evaluated at 10\%, 25\%, 50\%, 75\%, 100\%, and 200\% of this size, a 20-fold range chosen to cover data budgets below and above the reference, with the 200\% datasets collected from the same behavior policies.

\paragraph{Flat regime.}
Flat FQE's error plateaus across all $N$, confirming that $\Delta(\Pi)$ is invariant to dataset size. By contrast, IO-FQE-BE and IO-FQE-AC both bypass this bottleneck and exhibit consistent error reduction as $N$ increases, with IO-FQE-BE maintaining a lower error floor than IO-FQE-AC. The Monte Carlo approximation variance is only one component of IO-FQE-AC's error (Appendix~\ref{app:mc_samples}), and the difference in error floors is consistent with its remaining regression, coverage, and extrapolation terms.

\paragraph{Hierarchical regime.}
IO-FQE-BE improves initially as $N$ increases, but its error plateaus at larger dataset sizes; in Puzzle-$3\!\times\!3$, the error reaches its minimum at 75\% of the dataset and increases slightly beyond that point. This plateau reflects the limits of density-based correction under strict option commitment: additional data refines the density estimate around the narrow behavior trajectories but does not improve coverage over the action distributions of other options, setting a floor on the regression error.

IO-FQE-AC follows a more consistent descent. At small $N$, the Q-network is queried at actions sampled from $\pi_o$ that lie far from the sparse data trajectories, producing high variance. As $N$ increases, the data trajectories become sufficiently dense for the Q-network to interpolate stably even at actions outside the data support, and the MSE drops below that of IO-FQE-BE.

The dataset size at which IO-FQE-AC overtakes IO-FQE-BE depends on the degree of overlap among the intra-option action distributions. In AntMaze-L, the eight-dimensional action space is shared across locomotion options with moderate overlap, and IO-FQE-AC overtakes IO-FQE-BE between 50\% and 75\% of the dataset. In Puzzle-$3\!\times\!3$, the three manipulation primitives produce near-disjoint action distributions, and IO-FQE-AC outperforms IO-FQE-BE across all evaluated dataset sizes, with the performance gap widening to an order of magnitude at $N \ge 100\%$. 
\section{Related Work}
\label{sec:related_work}

Three lines of work bear on IO-FQE: flat off-policy evaluation, evaluation of hierarchical policies, and behavior density estimation for importance-weighted value learning.

Off-policy evaluation of Markovian target policies is well studied via importance sampling~\citep{precup2000eligibility, jiang2016doubly, liu2018breaking}, value-function regression~\citep{le2019batch, munos2008finite}, and hybrid estimators~\citep{nachum2019dualdice, uehara2020minimax}, with FQE performing strongly on continuous-action benchmarks~\citep{voloshin2019empirical, fu2021benchmarks, duan2020minimax}; none accommodate non-Markovian hierarchical targets.

The closest prior work, \citet{guo2017using}, applies per-decision importance sampling at the option level with incremental variance reduction (INCRIS). Both target $J(\Pi)$, but the approaches differ in three respects: \citet{guo2017using} is IS-based and inherits horizon-dependent variance, operates at semi-Markov option boundaries rather than the primitive time scale, and requires trajectories annotated with option identities and termination events. INCRIS does not address the variance scaling in continuous action spaces (Section~\ref{sec:experiments}). IO-FQE instead casts hierarchical OPE as value-function estimation in $\sS \times \sO$, operating at the primitive time scale and requiring no option annotations.

IO-FQE-BE follows the regression importance sampling framework~\citep{hanna2019importance, hanna2021importance}, and IO-FQE-AC avoids density estimation by conditioning on the primitive action. Further discussion of abstraction-based OPE~\citep{pavse2023scaling, pavse2023state, pavse2025stable, hao2024off, chaudhari2024abstract}, option discovery~\citep{bacon2017option, drappo2023option, drappo2024provably}, OPE under partial observability~\citep{uehara2023future}, and related variance-reduction techniques~\citep{rowland2020conditional, kallus2018policy, zhou2025demystifying} is in Appendix~\ref{app:related_work}.
\section{Conclusion}
\label{sec:discussion}

This paper recasts off-policy evaluation of committed option policies as a target-mismatch rather than an estimation problem: flat FQE consistently estimates $J(\pi_M)$, which differs from $J(\Pi)$ by a gap $\Delta(\Pi)$ that no amount of data closes. Lifting evaluation to the augmented state-option space $\sS \times \sO$ removes this mismatch without requiring option-annotated data, and IO-FQE-BE and IO-FQE-AC instantiate this lift through importance weighting and Monte Carlo marginalization, respectively.

IO-FQE assumes the target's option structure---the option set, intra-option policies, and termination functions---is given rather than recovered from data, and its per-transition cost scales linearly with $|\sO|$, suiting settings with a moderate number of structured options. Both variants further require every intra-option policy to be supported by the data distribution at every visited state. The empirical evaluation, finally, covers only two continuous-control simulation benchmarks. In these benchmarks, the flat and hierarchical data regimes expose different coverage conditions but do not constitute a controlled sweep over the divergence between the behavior and target policies. Likewise, the evaluated tasks, which use at most five options, do not isolate how the estimation error scales with $|\sO|$; beyond the linear per-transition cost, this statistical scaling may depend on per-option coverage. The two variants use MLPs of matched width ($[256, 256]$ hidden layers), although matched width does not imply equal approximation difficulty, because IO-FQE-AC represents the higher-dimensional function $Q(s, o, a)$; the comparison between the two variants may therefore be sensitive to network capacity.

There are four directions remain for future work. First, the assumption that the target's option structure is given motivates joint discovery of options and value from data; symmetrically, when data are collected by a hierarchical process with unobserved options, inferring option posteriors alongside the value function could capture the latent option mixture without fitting its marginal. Second, IO-FQE applies whenever the high-level policy selects among a finite library of low-level controllers with known intra-option policies and termination functions. The hierarchical systems that motivate this work use continuous skill embeddings~\citep{peng2022ase}, latent subgoals~\citep{park2023hiql}, or open-ended high-level commands~\citep{shi2025hi}, and evaluating them directly requires extending IO-FQE beyond a finite known option set. Third, a hybrid estimator that adapts between IO-FQE-BE and IO-FQE-AC across the state-action space could combine their strengths. Fourth, applying IO-FQE to evaluate high-level planners over shared skill libraries on real robotic systems---a setting in which offline logs typically lack skill annotations---would test the method in its native operating regime.

\begin{ack}
This work was conducted in the Prediction and Action Lab (PAL) at UW--Madison. PAL is supported by the National Science Foundation (IIS-2410981 and IIS-2551198) and by the Wisconsin Alumni Research Foundation.
\end{ack}

\clearpage
\bibliographystyle{abbrvnat}
\bibliography{ref} 

\clearpage
\appendix
\section{Extended Related Work}
\label{app:related_work}

This appendix expands the discussion in Section~\ref{sec:related_work}, situating IO-FQE within three areas: off-policy evaluation in flat MDPs, evaluation of hierarchical policies, and behavior density estimation for importance-weighted value-function learning.

\paragraph{Off-policy evaluation and the role of state representation.}
Off-policy evaluation (OPE) of a Markovian target policy has developed along three broad families: importance-sampling estimators that reweight observed returns by density ratios between target and behavior policies~\citep{precup2000eligibility, liu2018breaking, jiang2016doubly, thomas2016data}, value-function estimators that iteratively regress Bellman targets~\citep{le2019batch, munos2008finite, antos2008learning}, and hybrid approaches that combine elements of both~\citep{nachum2019dualdice, zhang2020gendice, uehara2020minimax}. Among these, fitted Q-evaluation (FQE)~\citep{le2019batch} performs well in empirical benchmarks on continuous-action offline data~\citep{voloshin2019empirical, fu2021benchmarks} and admits error analyses under linear~\citep{duan2020minimax} and nonparametric~\citep{wang2024fine} function approximation. A complementary line of work uses abstraction to improve OPE. Coarsening the state space lowers the variance of marginalized importance sampling ratios~\citep{pavse2023scaling} and, more generally, trades a bias term from loss of Markov structure against a variance term that decreases with abstraction~\citep{hao2024off, chaudhari2024abstract}; learning state-action representations from an OPE-tailored behavioral similarity metric improves the data efficiency of FQE~\citep{pavse2023state}, and bisimulation-based representations stabilize offline value function learning~\citep{pavse2025stable}. IO-FQE sits in a symmetric position within this program: abstraction coarsens the state space to reduce variance, and IO-FQE refines it from $\sS$ to $\sS \times \sO$ to eliminate a structural bias that arises when flat FQE is applied to a non-Markovian hierarchical target.

\paragraph{Off-policy evaluation of hierarchical policies.}
The options framework~\citep{sutton1999between, precup2000temporal} and the intra-option perspective~\citep{sutton1998intra} provide the formal and algorithmic foundations on which IO-FQE builds. The OPE literature, however, has engaged far less with hierarchical target policies than with option discovery or option learning~\citep{bacon2017option, drappo2023option, drappo2024provably}. The closest prior work is that of \citet{guo2017using}, who observe that hierarchical policies admit lower-variance IS estimators relative to flat trajectory-level reweighting, and introduce incremental importance sampling with covariance testing to prune statistically uninformative weights. Both \citet{guo2017using} and IO-FQE target the true hierarchical return $J(\Pi)$, but the two approaches differ in three structural respects: the estimator of \citet{guo2017using} is IS-based and inherits the horizon-dependent variance of trajectory reweighting; it operates at semi-Markov option boundaries rather than at the primitive time scale; and it requires trajectories annotated with option-initiation and termination events. The variance reduction relative to flat IS does not eliminate the fundamental scaling problem: in continuous action spaces, per-step action-density ratios compound across the trajectory, and the resulting variance remains prohibitive in long-horizon tasks (Section~\ref{sec:experiments}). IO-FQE instead casts hierarchical OPE as value-function estimation in the augmented space $\sS \times \sO$, operates at the primitive time scale, and requires no option annotations in the data. A separate line of work addresses non-Markovian target policies arising from partial observability~\citep{uehara2023future}; these methods confront non-Markovianity at the environment level, whereas IO-FQE addresses non-Markovianity that arises structurally from the hierarchical target policy in a fully observed MDP.

\paragraph{Behavior density estimation in importance-weighted evaluation.}
IO-FQE-BE (Section~\ref{subsec:iofqe_be}) corrects the mismatch between the intra-option policy $\pi_o$ and the data action density via importance weighting, placing it within the regression importance sampling (RIS) framework of \citet{hanna2019importance, hanna2021importance}. The RIS framework establishes that estimating the behavior policy from the evaluation dataset can reduce asymptotic variance relative to using the true behavior policy, with recent extensions to history-dependent behavior policies~\citep{zhou2025demystifying}. More broadly, conditioning IS weights on additional structure is a recurring variance-reduction strategy: \citet{rowland2020conditional} show that replacing an IS ratio by its conditional expectation given a sufficient functional is variance non-increasing, and IO-FQE-BE can be viewed as an instance of this principle with the option index as the conditioning variable. In continuous action spaces, however, accurate density estimation faces two difficulties: kernel-based estimators suffer bandwidth-sensitive variance~\citep{kallus2018policy}, and parametric density models are prone to misspecification as action dimensionality grows, particularly when the behavior distribution is a history-dependent mixture of intra-option policies. IO-FQE-AC (Section~\ref{subsec:iofqe_ac}) is motivated by these limitations: by conditioning on the primitive action in the value function, it sidesteps behavior density estimation entirely and replaces unbounded IS variance with bounded Monte Carlo variance that decreases as $O(1/M)$ in the number of policy samples.
\section{Algorithm Descriptions}
\label{app:algorithms}

This section provides complete pseudocode for the two IO-FQE estimation strategies presented in Section~\ref{sec:method}. Algorithm~\ref{alg:io_fqe_be} describes IO-FQE-BE, which requires a pre-estimated behavior action density. Algorithm~\ref{alg:io_fqe_ac} describes IO-FQE-AC, which replaces importance weighting with Monte Carlo marginalization.

\begin{algorithm}[h]
\caption{IO-FQE-BE}
\label{alg:io_fqe_be}
\begin{algorithmic}[1]
\INPUT Dataset $\mathcal{D} = \{(s_i, a_i, r_i, s'_i)\}_{i=1}^N$, hierarchical target policy $\Pi = (\mu, \sO)$ with intra-option policies $\{\pi_o\}_{o \in \sO}$ and termination functions $\{\beta_o\}_{o \in \sO}$, target network update rate $\tau$
\OUTPUT Option-value function $Q_\theta(s, o)$
\STATE Estimate $\hat{p}_\mathcal{D}(a \mid s)$ via maximum likelihood on $\mathcal{D}$
\STATE Initialize Q-network $\theta$ and target network $\bar{\theta} \leftarrow \theta$
\FOR{step $= 1$ \TO $N_{\text{steps}}$}
    \STATE Sample mini-batch $\mathcal{B} \subset \mathcal{D}$
    \FOR{each $(s, a, r, s') \in \mathcal{B}$}
        \STATE $V_{\bar{\theta}}(s') \leftarrow \sum_{o'} \mu(o' \mid s') \, Q_{\bar{\theta}}(s', o')$
        \FOR{each $o \in \sO$}
            \STATE $U_{\bar{\theta}}(s', o) \leftarrow (1 - \beta_o(s')) \, Q_{\bar{\theta}}(s', o) + \beta_o(s') \, V_{\bar{\theta}}(s')$
            \STATE $y_o \leftarrow r + \gamma \, U_{\bar{\theta}}(s', o)$
            \STATE $w_o \leftarrow \pi_o(a \mid s) \,/\, \hat{p}_{\mathcal{D}}(a \mid s)$
        \ENDFOR
    \ENDFOR
    \STATE Update $\theta$ by minimizing $\frac{1}{|\mathcal{B}|} \sum_{(s,a) \in \mathcal{B}} \sum_{o \in \sO} w_o \big( Q_\theta(s, o) - y_o \big)^2$
    \STATE Soft-update target network: $\bar{\theta} \leftarrow \tau \theta + (1 - \tau) \bar{\theta}$
\ENDFOR
\end{algorithmic}
\end{algorithm}

\paragraph{Weight stabilization.}
In continuous action spaces, the raw importance weights $w_o(s, a) = \pi_o(a \mid s) / \hat{p}_\mathcal{D}(a \mid s)$ can exhibit large variance when $\hat{p}_\mathcal{D}$ assigns low density to actions favored by $\pi_o$. In practice, two standard remedies are available, of which only the second is applied in the experiments. First, weight clipping truncates each weight at a maximum value $C_{\max}$, replacing $w_o$ with $\min(w_o,\, C_{\max})$~\cite{bottou2013counterfactual}. Second, self-normalization replaces the raw weights with $\tilde{w}_o = w_o / \sum_{o' \in \sO} w_{o'}$ within each transition, producing a weighted importance sampling variant~\cite{swaminathan2015self}. Both techniques introduce bias: clipping underweights actions in the tails of $\pi_o$, and self-normalization yields a consistent but biased estimator. The resulting bias-variance tradeoff is controlled by $C_{\max}$ (for clipping) or the effective sample size (for self-normalization). The Bellman consistency established in Proposition~\ref{prop:bellman_consistency_be} holds for the unmodified weights; with clipping or self-normalization, the population fixed point is in general only approximately equal to $Q^\Pi(s, o)$.

\begin{algorithm}[h]
\caption{IO-FQE-AC}
\label{alg:io_fqe_ac}
\begin{algorithmic}[1]
\INPUT Dataset $\mathcal{D} = \{(s_i, a_i, r_i, s'_i)\}_{i=1}^N$, hierarchical target policy $\Pi = (\mu, \sO)$ with intra-option policies $\{\pi_o\}_{o \in \sO}$ and termination functions $\{\beta_o\}_{o \in \sO}$, target network update rate $\tau$, number of MC samples $M$
\OUTPUT Action-conditioned option-value function $Q_\theta(s, o, a)$
\STATE Initialize Q-network $\theta$ and target network $\bar{\theta} \leftarrow \theta$
\FOR{step $= 1$ \TO $N_{\text{steps}}$}
    \STATE Sample mini-batch $\mathcal{B} \subset \mathcal{D}$
    \FOR{each $(s, a, r, s') \in \mathcal{B}$}
        \FOR{each $o \in \sO$}
            \STATE Sample $\{a'_m\}_{m=1}^M \sim \pi_o(\cdot \mid s')$ \hfill $\triangleright$ MC marginalization
            \STATE $\hat{Q}_{\bar{\theta}}(s', o) \leftarrow \frac{1}{M} \sum_{m=1}^{M} Q_{\bar{\theta}}(s', o, a'_m)$
        \ENDFOR
        \STATE $V_{\bar{\theta}}(s') \leftarrow \sum_{o'} \mu(o' \mid s') \, \hat{Q}_{\bar{\theta}}(s', o')$
        \FOR{each $o \in \sO$}
            \STATE $U_{\bar{\theta}}(s', o) \leftarrow (1 - \beta_o(s')) \, \hat{Q}_{\bar{\theta}}(s', o) + \beta_o(s') \, V_{\bar{\theta}}(s')$
            \STATE $y_o \leftarrow r + \gamma \, U_{\bar{\theta}}(s', o)$
        \ENDFOR
    \ENDFOR
    \STATE Update $\theta$ by minimizing $\frac{1}{|\mathcal{B}|} \sum_{(s,a) \in \mathcal{B}} \sum_{o \in \sO} \big( Q_\theta(s, o, a) - y_o \big)^2$
    \STATE Soft-update target network: $\bar{\theta} \leftarrow \tau \theta + (1 - \tau) \bar{\theta}$
\ENDFOR
\end{algorithmic}
\end{algorithm}
\section{Theoretical Analysis}
\label{app:theory}

This section provides the proof of Proposition~\ref{prop:irreducible_bias} deferred from Section~\ref{sec:problem_formulation}, establishes that the iterative procedure underlying IO-FQE-BE and IO-FQE-AC converges to the correct intra-option value functions at the population level (Section~\ref{app:bellman_consistency}), compares the assumptions and error pathways of both methods against each other and provides finite-sample error bounds (Section~\ref{app:error_bounds}).

The analysis proceeds in two stages that mirror the algorithmic structure. At each iteration, FQE fixes a target function $\bar{Q}$ and solves a regression problem against targets computed from $\bar{Q}$; the solution defines a Bellman operator $\mathcal{T}$ via $\mathcal{T}\bar{Q} = \arg\min_Q \mathcal{L}(Q;\bar{Q})$. The fixed point of $\mathcal{T}$ is then identified as the true value function through a contraction argument. This two-stage decomposition, regression against fixed targets followed by fixed-point iteration, avoids conflating the optimization objective with the Bellman equation and corresponds directly to the target-network updates in Algorithms~\ref{alg:io_fqe_be} and~\ref{alg:io_fqe_ac}. 


\begin{proof}[Proof of Proposition~\ref{prop:irreducible_bias}]
\label{app:proofs_sec2}
The upper bound follows from the triangle inequality applied to the decomposition $\hat{J}_{\mathrm{FQE}}(\pi_M; \mathcal{D}) - J(\Pi) = \big(\hat{J}_{\mathrm{FQE}}(\pi_M; \mathcal{D}) - J(\pi_M)\big) + \big(J(\pi_M) - J(\Pi)\big)$, which yields $\big|\hat{J}_{\mathrm{FQE}}(\pi_M; \mathcal{D}) - J(\Pi)\big| \le \epsilon_{\mathrm{est}}(\mathcal{D}) + \Delta(\Pi)$. The lower bound follows from the triangle inequality applied to the decomposition $J(\pi_M) - J(\Pi) = \big(J(\pi_M) - \hat{J}_{\mathrm{FQE}}(\pi_M; \mathcal{D})\big) + \big(\hat{J}_{\mathrm{FQE}}(\pi_M; \mathcal{D}) - J(\Pi)\big)$, which yields $\Delta(\Pi) \le \epsilon_{\mathrm{est}}(\mathcal{D}) + \big|\hat{J}_{\mathrm{FQE}}(\pi_M; \mathcal{D}) - J(\Pi)\big|$. Rearranging the second inequality and combining it with the first gives Eq.~\eqref{eq:bias_decomp}.
\end{proof}

\begin{proof}[Proof of Corollary~\ref{cor:irreducibility}]
Let $\mathcal{D}_N$ denote a dataset of size $N$, and write $E_N \triangleq \big|\hat{J}_{\mathrm{FQE}}(\pi_M; \mathcal{D}_N) - J(\Pi)\big|$ and $\epsilon_N \triangleq \epsilon_{\mathrm{est}}(\mathcal{D}_N)$. Instantiating Eq.~\eqref{eq:bias_decomp} at $\mathcal{D} = \mathcal{D}_N$ yields $|E_N - \Delta(\Pi)| \le \epsilon_N$ for every realization of $\mathcal{D}_N$. On the probability-one event on which $\epsilon_N \to 0$, this bound gives $E_N \to \Delta(\Pi)$; hence $E_N \to \Delta(\Pi)$ with probability $1$.
\end{proof}

Throughout, the shorthand $U_{\bar{Q}}(s, o) \triangleq (1 - \beta_o(s))\, \bar{Q}(s, o) + \beta_o(s)\, V_{\bar{Q}}(s)$ is used, where $V_{\bar{Q}}(s) \triangleq \sum_{o'} \mu(o' \mid s)\, \bar{Q}(s, o')$.

\subsection{Bellman Consistency}
\label{app:bellman_consistency}

\subsubsection{IO-FQE-BE}

\begin{assumption}[Coverage for IO-FQE-BE]
\label{assump:coverage_be}
For all $o \in \sO$ and all $s$ in the support of the data distribution, $\pi_o(\cdot \mid s) \ll p_\mathcal{D}(\cdot \mid s)$, and the importance ratio is bounded: $\sup_{s, a, o} \frac{\pi_o(a \mid s)}{p_\mathcal{D}(a \mid s)} \leq C$ for some finite constant $C > 0$. Rewards are bounded: $|R(s, a)| \leq R_{\max}$ for all $(s, a)$.
\end{assumption}

\begin{proposition}[Bellman Consistency of IO-FQE-BE]
\label{prop:bellman_consistency_be}
Under Assumption~\ref{assump:coverage_be}, suppose $w_o(s, a) = \pi_o(a \mid s) / p_\mathcal{D}(a \mid s)$ uses the true behavior action density. For any fixed bounded measurable function $\bar{Q}: \sS \times \sO \to \mathbb{R}$, define the regression objective
\begin{equation}
    \label{eq:pop_loss_be}
    \mathcal{L}_{\mathrm{BE}}(Q;\, \bar{Q}) = \mathbb{E}_{s \sim d_\mathcal{D},\, a \sim p_\mathcal{D}(\cdot \mid s),\, s' \sim P(\cdot \mid s, a)} \left[ \sum_{o \in \sO} w_o(s, a) \left( Q(s, o) - y_o(s, a, s') \right)^2 \right],
\end{equation}
where $y_o(s, a, s') = R(s, a) + \gamma\, U_{\bar{Q}}(s', o)$ is computed from $\bar{Q}$ and held fixed during optimization over $Q$. Then:
\begin{enumerate}
    \item[(i)] Over the space of all bounded measurable functions $Q: \sS \times \sO \to \mathbb{R}$, the minimizer of $\mathcal{L}_{\mathrm{BE}}(Q;\, \bar{Q})$ satisfies the option-conditioned intra-option Bellman equation with $\bar{Q}$ as the bootstrap function: for all $s$ with $d_\mathcal{D}(s) > 0$ and all $o \in \sO$,
    \begin{equation}
        Q^*(s, o) = \mathbb{E}_{a \sim \pi_o(\cdot \mid s)} \!\Big[ R(s, a) + \gamma \, \mathbb{E}_{s' \sim P(\cdot \mid s, a)}\!\big[ U_{\bar{Q}}(s', o) \big] \Big] \triangleq (\mathcal{T}_{\mathrm{BE}}\, \bar{Q})(s, o).
    \end{equation}
    \item[(ii)] The operator $\mathcal{T}_{\mathrm{BE}}$ is a $\gamma$-contraction in the sup-norm on the space of bounded measurable functions on $\sS \times \sO$. By the Banach fixed point theorem, $\mathcal{T}_{\mathrm{BE}}$ has a unique fixed point, which is $Q^\Pi(s, o)$.
\end{enumerate}
\end{proposition}

\begin{proof}
\textup{(i)} Since $\bar{Q}$ is fixed, $y_o(s, a, s')$ does not depend on $Q$, and $\mathcal{L}_{\mathrm{BE}}(Q;\, \bar{Q})$ is a weighted least-squares problem in $Q$. The loss decomposes across states: for each $s$ with $d_\mathcal{D}(s) > 0$ and each $o \in \sO$, the first-order condition with respect to $Q(s, o)$ yields
\begin{equation}
    \mathbb{E}_{a \sim p_\mathcal{D}(\cdot \mid s),\, s' \sim P(\cdot \mid s, a)} \Big[ w_o(s, a) \big( Q(s, o) - y_o(s, a, s') \big) \Big] = 0.
\end{equation}
Since $Q(s, o)$ does not depend on $a$, separating terms gives
\begin{equation}
    Q(s, o) \cdot \mathbb{E}_{a \sim p_\mathcal{D}}[w_o(s, a)] = \mathbb{E}_{a \sim p_\mathcal{D}} \Big[ w_o(s, a) \cdot \mathbb{E}_{s' \mid s, a}[y_o(s, a, s')] \Big].
\end{equation}
The left-hand side simplifies via the normalization identity:
\begin{equation}
    \mathbb{E}_{a \sim p_\mathcal{D}(\cdot \mid s)}[w_o(s, a)] = \int_\sA \frac{\pi_o(a \mid s)}{p_\mathcal{D}(a \mid s)} \, p_\mathcal{D}(a \mid s) \, da = \int_\sA \pi_o(a \mid s) \, da = 1,
\end{equation}
where the cancellation is valid because $p_\mathcal{D}(a \mid s) > 0$ on the support of $\pi_o(\cdot \mid s)$ by Assumption~\ref{assump:coverage_be}. The right-hand side undergoes the same change of measure:
\begin{align}
    \mathbb{E}_{a \sim p_\mathcal{D}} \Big[ w_o(s, a) \cdot \mathbb{E}_{s' \mid s, a}[y_o] \Big] &= \int_\sA \pi_o(a \mid s) \, \mathbb{E}_{s' \mid s, a}\!\big[R(s,a) + \gamma\, U_{\bar{Q}}(s', o)\big] \, da \\
    &= \mathbb{E}_{a \sim \pi_o(\cdot \mid s)} \!\Big[ R(s, a) + \gamma \, \mathbb{E}_{s' \sim P(\cdot \mid s, a)}\!\big[ U_{\bar{Q}}(s', o) \big] \Big] = (\mathcal{T}_{\mathrm{BE}}\, \bar{Q})(s, o).
\end{align}

\textup{(ii)} For any two bounded measurable functions $\bar{Q}_1, \bar{Q}_2$,
\begin{align}
    |(\mathcal{T}_{\mathrm{BE}}\, \bar{Q}_1)(s, o) - (\mathcal{T}_{\mathrm{BE}}\, \bar{Q}_2)(s, o)| &= \gamma \, \big| \mathbb{E}_{a \sim \pi_o,\, s' \sim P}\!\big[ U_{\bar{Q}_1}(s', o) - U_{\bar{Q}_2}(s', o) \big] \big| \\
    &\leq \gamma \, \| \bar{Q}_1 - \bar{Q}_2 \|_\infty,
\end{align}
where the inequality follows from the definition of $U_{\bar{Q}}$ as a convex combination involving $\bar{Q}$ and $V_{\bar{Q}}$, both of which are linear in $\bar{Q}$. Hence $\mathcal{T}_{\mathrm{BE}}$ is a $\gamma$-contraction. By the Banach fixed point theorem, a unique fixed point exists. Since $Q^\Pi$ satisfies $Q^\Pi = \mathcal{T}_{\mathrm{BE}}\, Q^\Pi$ (the intra-option Bellman equation, Eq.~\eqref{eq:bellman_so}), this fixed point is $Q^\Pi$.
\end{proof}

\begin{remark}[Weight placement: outside the squared error vs.\ inside the regression target]
\label{rem:weight_placement}
A natural alternative to $\mathcal{L}_{\mathrm{BE}}$ corrects the regression target directly: replacing the conditional expectation $\mathbb{E}_{a \sim p_\mathcal{D}}[y_o \mid s]$ with its importance-corrected counterpart $\mathbb{E}_{a \sim p_\mathcal{D}}[w_o(s,a)\,y_o \mid s] = \mathbb{E}_{a \sim \pi_o}[y_o \mid s]$, and regressing $Q_\theta(s,o)$ against this corrected target under unweighted mean-squared error. The two formulations induce the same population Bellman backup. Conditioning on $s$ and exploiting that $Q_\theta(s,o)$ does not depend on $a$, the inner expectation of $\mathcal{L}_{\mathrm{BE}}$ expands as
\begin{align*}
    \mathbb{E}_{a \sim p_\mathcal{D}}\bigl[w_o(Q_\theta - y_o)^2 \,\big|\, s\bigr]
    &= Q_\theta(s,o)^2\,\mathbb{E}_{a \sim p_\mathcal{D}}[w_o \mid s] - 2\,Q_\theta(s,o)\,\mathbb{E}_{a \sim p_\mathcal{D}}[w_o\,y_o \mid s] + \mathbb{E}_{a \sim p_\mathcal{D}}[w_o\,y_o^2 \mid s] \\
    &= Q_\theta(s,o)^2 - 2\,Q_\theta(s,o)\,\mathbb{E}_{a \sim \pi_o}[y_o \mid s] + \mathbb{E}_{a \sim p_\mathcal{D}}[w_o\,y_o^2 \mid s],
\end{align*}
where the first equality uses the linearity of expectation and that $Q_\theta(s,o)$ factors out of the action expectation, and the second uses the normalization identity $\mathbb{E}_{a \sim p_\mathcal{D}}[w_o(s,a) \mid s] = 1$ and the change of measure $\mathbb{E}_{a \sim p_\mathcal{D}}[w_o\,y_o \mid s] = \mathbb{E}_{a \sim \pi_o}[y_o \mid s]$. The right-hand side is a quadratic in $Q_\theta(s,o)$ whose unique minimizer over the unrestricted function class at each $(s,o)$ with $d_\mathcal{D}(s) > 0$ is
\begin{equation*}
    Q^*(s,o) = \mathbb{E}_{a \sim \pi_o}[y_o \mid s],
\end{equation*}
which is precisely the importance-corrected conditional target and equals $(\mathcal{T}_{\mathrm{BE}}\,\bar{Q})(s,o)$ from Proposition~\ref{prop:bellman_consistency_be}.

The weighted-residual formulation $\mathcal{L}_{\mathrm{BE}}$ implements this Bellman backup through weighted least squares; the weights themselves use the behavior density estimated from $\mathcal{D}$, following regression importance sampling~\citep{hanna2019importance,hanna2021importance}. The equivalence between the two formulations relies on the prediction $Q_\theta(s,o)$ being independent of the primitive action $a$. When the prediction conditions on $a$---as for the action-conditioned $Q_\theta(s,o,a)$ in IO-FQE-AC (Section~\ref{subsec:iofqe_ac})---the prediction no longer factors out of the action expectation, and weighted-residual regression and target correction induce different population minimizers; this asymmetry partly motivates the alternative Monte Carlo strategy adopted in IO-FQE-AC.
\end{remark}

\subsubsection{IO-FQE-AC}

\begin{assumption}[Coverage for IO-FQE-AC]
\label{assump:coverage_ac}
For all $o \in \sO$ and all $s$ in the support of the data distribution, $\pi_o(\cdot \mid s) \ll p_\mathcal{D}(\cdot \mid s)$ (coverage). Rewards are bounded: $|R(s, a)| \leq R_{\max}$ for all $(s, a)$.
\end{assumption}

\begin{proposition}[Bellman Consistency of IO-FQE-AC]
\label{prop:bellman_consistency_ac}
Under Assumption~\ref{assump:coverage_ac}, for any fixed bounded measurable function $\bar{Q}: \sS \times \sO \times \sA \to \mathbb{R}$, define the regression objective
\begin{equation}
    \label{eq:pop_loss_ac}
    \mathcal{L}_{\mathrm{AC}}(Q;\, \bar{Q}) = \mathbb{E}_{s \sim d_\mathcal{D},\, a \sim p_\mathcal{D}(\cdot \mid s),\, s' \sim P(\cdot \mid s, a)} \left[ \sum_{o \in \sO} \left( Q(s, o, a) - y_o(s, a, s') \right)^2 \right],
\end{equation}
where $y_o(s, a, s') = R(s, a) + \gamma\, U_{\bar{Q}}(s', o)$ is computed from $\bar{Q}$ (with $\bar{Q}(s, o) \triangleq \mathbb{E}_{a \sim \pi_o(\cdot \mid s)}[\bar{Q}(s, o, a)]$) and held fixed during optimization over $Q$. Then:
\begin{enumerate}
    \item[(i)] Over the space of all bounded measurable functions $Q: \sS \times \sO \times \sA \to \mathbb{R}$, the minimizer of $\mathcal{L}_{\mathrm{AC}}(Q;\, \bar{Q})$ satisfies the action-conditioned intra-option Bellman equation with $\bar{Q}$ as the bootstrap function: for all $(s, a)$ with $d_\mathcal{D}(s) \cdot p_\mathcal{D}(a \mid s) > 0$ and all $o \in \sO$,
    \begin{equation}
        Q^*(s, o, a) = R(s, a) + \gamma \, \mathbb{E}_{s' \sim P(\cdot \mid s, a)}\!\big[ U_{\bar{Q}}(s', o) \big] \triangleq (\mathcal{T}_{\mathrm{AC}}\, \bar{Q})(s, o, a).
    \end{equation}
    \item[(ii)] The operator $\mathcal{T}_{\mathrm{AC}}$ is a $\gamma$-contraction in the sup-norm on the space of bounded measurable functions on $\sS \times \sO \times \sA$. By the Banach fixed point theorem, $\mathcal{T}_{\mathrm{AC}}$ has a unique fixed point, which is $Q^\Pi(s, o, a)$.
\end{enumerate}
\end{proposition}

\begin{proof}
\textup{(i)} Since $\bar{Q}$ is fixed, $y_o(s, a, s')$ does not depend on $Q$, and $\mathcal{L}_{\mathrm{AC}}(Q;\, \bar{Q})$ is an unweighted least-squares problem in $Q$. For each $(s, a)$ with $d_\mathcal{D}(s) \cdot p_\mathcal{D}(a \mid s) > 0$ and each $o \in \sO$, the minimizer over $Q(s, o, a)$ is the conditional expectation of the target:
\begin{equation}
    Q^*(s, o, a) = \mathbb{E}_{s' \sim P(\cdot \mid s, a)}\!\big[y_o(s, a, s')\big] = R(s, a) + \gamma \, \mathbb{E}_{s' \sim P(\cdot \mid s, a)}\!\big[ U_{\bar{Q}}(s', o) \big],
\end{equation}
which is the action-conditioned intra-option Bellman equation (Eq.~\eqref{eq:bellman_soa}) with $\bar{Q}$ as the bootstrap function. The behavior action density $p_\mathcal{D}(a \mid s)$ determines only the weighting of different $(s, a)$ pairs in the loss; the conditional expectation $\mathbb{E}[y_o \mid s, a]$ that defines $Q^*$ does not depend on $p_\mathcal{D}$.

The coverage condition $\pi_o(\cdot \mid s) \ll p_\mathcal{D}(\cdot \mid s)$ in Assumption~\ref{assump:coverage_ac} is required for the operator $\mathcal{T}_{\mathrm{AC}}$ to produce meaningful iterates: $U_{\bar{Q}}(s', o)$ depends on $\bar{Q}(s', o) = \mathbb{E}_{a \sim \pi_o}[\bar{Q}(s', o, a)]$, which requires $\bar{Q}(s', o, a)$ to be constrained by the regression at all $a$ in the support of $\pi_o(\cdot \mid s')$. Since the regression pins down $Q^*(s, o, a)$ only where $p_\mathcal{D}(a \mid s) > 0$, the coverage condition ensures that these points include the entire support of $\pi_o$. If the coverage condition fails, $Q^*(s', o, a)$ is unconstrained on portions of the support of $\pi_o(\cdot \mid s')$, the marginalization $\bar{Q}(s', o)$ is determined by arbitrary function values rather than by data-grounded regression, and the contraction property of $\mathcal{T}_{\mathrm{AC}}$ over the state-action distribution induced by $\Pi$ breaks down. 

\textup{(ii)} For any two bounded measurable functions $\bar{Q}_1, \bar{Q}_2$,
\begin{align}
    |(\mathcal{T}_{\mathrm{AC}}\, \bar{Q}_1)(s, o, a) - (\mathcal{T}_{\mathrm{AC}}\, \bar{Q}_2)(s, o, a)| &= \gamma \, \big| \mathbb{E}_{s' \sim P}\!\big[ U_{\bar{Q}_1}(s', o) - U_{\bar{Q}_2}(s', o) \big] \big| \\
    &\leq \gamma \, \| \bar{Q}_1 - \bar{Q}_2 \|_\infty,
\end{align}
where the inequality follows from $U_{\bar{Q}}$ being linear in $\bar{Q}$. Hence $\mathcal{T}_{\mathrm{AC}}$ is a $\gamma$-contraction. By the Banach fixed point theorem, a unique fixed point exists. Since $Q^\Pi$ satisfies $Q^\Pi = \mathcal{T}_{\mathrm{AC}}\, Q^\Pi$ (the action-conditioned intra-option Bellman equation, Eq.~\eqref{eq:bellman_soa}), this fixed point is $Q^\Pi$.
\end{proof}

\begin{remark}
\label{rmk:ac_vs_flat}
The regression step of IO-FQE-AC recovers the action-conditioned Bellman equation as a conditional expectation, mirroring the standard flat FQE argument. The technical distinction lies in the bootstrapping step: recovering $Q(s, o)$ from $Q(s, o, a)$ via marginalization over $\pi_o$ requires the regression to have constrained $Q$ on the support of $\pi_o$, a requirement enforced by Assumption~\ref{assump:coverage_ac}. In flat FQE, the analogous marginalization $V^\pi(s') = \mathbb{E}_{a' \sim \pi}[Q(s', a')]$ involves the same function class and does not introduce a separate coverage requirement beyond the concentrability coefficient. The comparison of the two error bounds in Section~\ref{app:error_bounds} elaborates on this distinction.
\end{remark}

\subsection{Error Propagation}
\label{app:error_bounds}

This section provides finite-sample error bounds for IO-FQE-BE and IO-FQE-AC. Both results follow the standard FQE error propagation template, bounding the distance between the learned $Q_\theta$ and $Q^\Pi$ in terms of per-iteration regression error amplified by the effective horizon, but differ in the composition of the per-iteration error.

\subsubsection{IO-FQE-BE}

\begin{assumption}[Finite-sample conditions for IO-FQE-BE]
\label{assump:finite_be}
In addition to Assumption~\ref{assump:coverage_be}: (a) the function class $\mathcal{F}_{\mathrm{BE}} \subset \{Q: \sS \times \sO \to \mathbb{R}\}$ used to represent $Q_\theta$ satisfies realizability ($Q^\Pi \in \mathcal{F}_{\mathrm{BE}}$) and Bellman completeness ($\mathcal{T}_{\mathrm{BE}} Q \in \mathcal{F}_{\mathrm{BE}}$ for all $Q \in \mathcal{F}_{\mathrm{BE}}$); (b) the density estimator $\hat{p}_\mathcal{D}$ satisfies $\sup_{s,a} |w_o(s,a) - \hat{w}_o(s,a)| \leq \epsilon_p$ where $\hat{w}_o(s,a) = \pi_o(a|s)/\hat{p}_\mathcal{D}(a|s)$.
\end{assumption}

Define the state-level concentrability coefficient
\begin{equation}
    \label{eq:concentrability_state}
    C_{\mathrm{state}} \triangleq \sup_{s, o} \frac{d^\Pi(s, o)}{d_\mathcal{D}(s)},
\end{equation}
where $d^\Pi(s, o)$ is the normalized discounted state-option visitation distribution induced by $\Pi$ and $d_\mathcal{D}(s)$ is the state marginal of the data distribution. This coefficient measures the state-level distributional shift between the occupancy of $\Pi$ and the data distribution.

\begin{theorem}[Error Bound for IO-FQE-BE]
\label{thm:error_be}
Under Assumptions~\ref{assump:coverage_be} and~\ref{assump:finite_be}, suppose that each weighted regression step satisfies
\begin{equation}
    \|Q_k - \hat{\mathcal{T}}_{\mathrm{BE}} Q_{k-1}\|_{2, d_\mathcal{D}} \leq \sqrt{C \cdot \epsilon_{\mathrm{stat}}(N, \mathcal{F}_{\mathrm{BE}})},
\end{equation}
and that the density estimation error is bounded by $\sup_{s,a,o} |w_o(s,a) - \hat{w}_o(s,a)| \leq \epsilon_p < 1$. Then the output $Q_K$ after $K$ iterations satisfies
\begin{equation}
    \|Q_K - Q^\Pi\|_{2, d^\Pi} \leq 2\gamma^K \frac{R_{\max}}{1 - \gamma} + \frac{1}{1 - \sqrt{\gamma}} \sqrt{C_{\mathrm{state}} \cdot C \cdot \epsilon_{\mathrm{stat}}(N, \mathcal{F}_{\mathrm{BE}})} + \frac{2 \epsilon_p R_{\max}}{(1 - \epsilon_p)(1 - \gamma)^2}.
\end{equation}
Consequently, by absorbing the constants into the asymptotic notation, the bound simplifies to:
\begin{equation}
    \|Q_K - Q^\Pi\|_{2, d^\Pi} \lesssim \gamma^K \frac{R_{\max}}{1 - \gamma} + \frac{1}{1 - \gamma} \underbrace{\sqrt{C_{\mathrm{state}} \cdot C \cdot \epsilon_{\mathrm{stat}}(N, \mathcal{F}_{\mathrm{BE}})}}_{\text{extrapolated weighted regression error}} + \frac{1}{(1 - \gamma)^2} \underbrace{\epsilon_p \cdot R_{\max}}_{\text{density estimation bias}}.
\end{equation}
\end{theorem}

\begin{proof}
The proof relies on the approximate policy iteration error propagation framework \citep{munos2005error, farahmand2010error}, adapted to the augmented state-option space $\sS \times \sO$.

Let $Q_k$ denote the learned value function at iteration $k$, and let $\mathcal{T}_{\mathrm{BE}}$ be the exact population Bellman operator defined in Proposition~\ref{prop:bellman_consistency_be}, which satisfies $Q^\Pi = \mathcal{T}_{\mathrm{BE}} Q^\Pi$. At iteration $k$, the algorithm uses the empirical operator $\hat{\mathcal{T}}_{\mathrm{BE}}$ defined by the estimated importance weights $\hat{w}_o$. 

\textbf{Step 1: Per-iteration error decomposition and weight perturbation.} 
Define the point-wise Bellman residual $e_k(s, o) \triangleq Q_k(s, o) - (\mathcal{T}_{\mathrm{BE}} Q_{k-1})(s, o)$. We decompose this residual into a statistical regression error $\epsilon_k$ and a weight perturbation error $\delta_k$:
\begin{equation}
    e_k(s, o) = \underbrace{Q_k(s, o) - (\hat{\mathcal{T}}_{\mathrm{BE}} Q_{k-1})(s, o)}_{\triangleq \, \epsilon_k(s, o)} \;+\; \underbrace{(\hat{\mathcal{T}}_{\mathrm{BE}} Q_{k-1})(s, o) - (\mathcal{T}_{\mathrm{BE}} Q_{k-1})(s, o)}_{\triangleq \, \delta_k(s, o)}.
\end{equation}
By assumption, the regression error satisfies $\|\epsilon_k\|_{2, d_\mathcal{D}} \leq \sqrt{C \cdot \epsilon_{\mathrm{stat}}(N, \mathcal{F}_{\mathrm{BE}})}$.

We now bound the perturbation $\delta_k$. The target is $\bar{y}_o(s, a) = R(s, a) + \gamma U_{Q_{k-1}}(s', o)$, bounded by $R_{\max}/(1-\gamma)$. The population minimizer under perturbed weights $\hat{w}_o$ follows from the first-order condition:
\begin{equation}
    (\hat{\mathcal{T}}_{\mathrm{BE}} Q_{k-1})(s, o) = \frac{\mathbb{E}_{a \sim p_\mathcal{D}}[\hat{w}_o(s,a) \bar{y}_o(s,a) \mid s]}{\mathbb{E}_{a \sim p_\mathcal{D}}[\hat{w}_o(s,a) \mid s]}.
\end{equation}
Let $\hat{w}_o = w_o + \eta$, where $|\eta(s,a)| \leq \epsilon_p$. Let $Q^*_{\mathrm{true}} \triangleq (\mathcal{T}_{\mathrm{BE}} Q_{k-1})(s,o) = \mathbb{E}_{a \sim \pi_o}[\bar{y}_o \mid s]$. Using $\mathbb{E}_{p_\mathcal{D}}[w_o \mid s] = 1$, we expand the error:
\begin{align}
    \delta_k(s, o) &= \frac{\mathbb{E}_{p_\mathcal{D}}[(w_o + \eta)\bar{y}_o \mid s]}{1 + \mathbb{E}_{p_\mathcal{D}}[\eta \mid s]} - Q^*_{\mathrm{true}} = \frac{\mathbb{E}_{p_\mathcal{D}}[\eta (\bar{y}_o - Q^*_{\mathrm{true}}) \mid s]}{1 + \mathbb{E}_{p_\mathcal{D}}[\eta \mid s]}.
\end{align}
Since $|\bar{y}_o - Q^*_{\mathrm{true}}| \leq 2R_{\max}/(1-\gamma)$ and the denominator is lower bounded by $1 - \epsilon_p > 0$, we obtain:
\begin{equation}
    \|\delta_k\|_{\infty} \leq \frac{2 \epsilon_p R_{\max}}{(1 - \epsilon_p)(1 - \gamma)} \triangleq \delta_{\max}.
\end{equation}

\textbf{Step 2: Operator unrolling.}
Unlike the $L_\infty$ norm, the Bellman operator is not generally a $\gamma$-contraction in the $L_2$ norm. Following \citet{farahmand2010error}, we unroll the error using the linear transition operator of the target policy. For any function $f$, define the intra-option transition operator $P^\Pi$ on $\sS \times \sO$ as $(P^\Pi f)(s, o) \triangleq \mathbb{E}_{a \sim \pi_o, s' \sim P}[U_f(s', o)]$. The exact Bellman iteration satisfies:
\begin{equation}
    Q_k - Q^\Pi = e_k + \gamma P^\Pi (Q_{k-1} - Q^\Pi).
\end{equation}
Unrolling this recursion over $K$ iterations, and taking the absolute value pointwise (since $P^\Pi$ is a monotone kernel with non-negative entries), we obtain:
\begin{equation}
    |Q_K - Q^\Pi| \leq \sum_{k=1}^K \gamma^{K-k} (P^\Pi)^{K-k} |e_k| \;+\; \gamma^K (P^\Pi)^K |Q_0 - Q^\Pi|.
\end{equation}

\textbf{Step 3: Error separation, norm conversion, and change of measure.}
Applying the triangle inequality to the pointwise error $|e_k| \leq |\epsilon_k| + |\delta_k|$, and using the linearity and monotonicity of $(P^\Pi)^{K-k}$, we separate the unrolling into two distinct pipelines:
\begin{equation}
    |Q_K - Q^\Pi| \leq \sum_{k=1}^K \gamma^{K-k} (P^\Pi)^{K-k} |\epsilon_k| \;+\; \sum_{k=1}^K \gamma^{K-k} (P^\Pi)^{K-k} |\delta_k| \;+\; \gamma^K (P^\Pi)^K |Q_0 - Q^\Pi|.
\end{equation}
Taking the $L_2(d^\Pi)$ norm on both sides, we bound each of the three terms individually.

For the initial error term, since value functions are uniformly bounded by $R_{\max}/(1-\gamma)$, the difference $\|Q_0 - Q^\Pi\|_\infty \leq 2R_{\max}/(1-\gamma)$, and the $L_2(d^\Pi)$ norm is bounded by $2\gamma^K R_{\max}/(1-\gamma)$.

For the weight perturbation term $\delta_k$, since $P^\Pi$ is a stochastic transition kernel, it satisfies $\|(P^\Pi)^m f\|_\infty \leq \|f\|_\infty$ for any function $f$. Using the uniform bound $\|\delta_k\|_\infty \leq \delta_{\max}$:
\begin{equation}
    \Big\| \sum_{k=1}^K \gamma^{K-k} (P^\Pi)^{K-k} |\delta_k| \Big\|_{2, d^\Pi} \leq \sum_{k=1}^K \gamma^{K-k} \|\delta_k\|_\infty \leq \frac{\delta_{\max}}{1 - \gamma} = \frac{2 \epsilon_p R_{\max}}{(1 - \epsilon_p)(1 - \gamma)^2}.
\end{equation}

For the statistical regression error term $\epsilon_k$, we apply the change of measure. By Jensen's inequality for Markov kernels, $[(P^\Pi)^m |\epsilon_k|]^2 \leq (P^\Pi)^m (\epsilon_k^2)$. Using the property of the discounted visitation distribution $d^\Pi (P^\Pi)^m \leq \gamma^{-m} d^\Pi$, and introducing the state-level concentrability coefficient $C_{\mathrm{state}} = \sup \frac{d^\Pi}{d_\mathcal{D}}$:
\begin{align}
    \big\| (P^\Pi)^m |\epsilon_k| \big\|_{2, d^\Pi}^2 \leq \int_{\sS \times \sO} d^\Pi \cdot (P^\Pi)^m (\epsilon_k^2) &\leq \gamma^{-m} \int_{\sS \times \sO} d^\Pi \epsilon_k^2 \nonumber \\
    &\leq \gamma^{-m} C_{\mathrm{state}} \int_{\sS \times \sO} d_\mathcal{D} \epsilon_k^2 = \gamma^{-m} C_{\mathrm{state}} \|\epsilon_k\|_{2, d_\mathcal{D}}^2.
\end{align}
Taking the square root and substituting $m = K-k$, the mismatched $\gamma$ factors cancel:
\begin{align}
    \sum_{k=1}^K \gamma^{K-k} \big\| (P^\Pi)^{K-k} |\epsilon_k| \big\|_{2, d^\Pi} &\leq \sum_{k=1}^K \gamma^{(K-k)/2} \sqrt{C_{\mathrm{state}}} \|\epsilon_k\|_{2, d_\mathcal{D}} \nonumber \\
    &\leq \frac{1}{1 - \sqrt{\gamma}} \sqrt{C_{\mathrm{state}}} \max_{1 \le k \le K} \|\epsilon_k\|_{2, d_\mathcal{D}}.
\end{align}
Substituting $\|\epsilon_k\|_{2, d_\mathcal{D}} \leq \sqrt{C \cdot \epsilon_{\mathrm{stat}}(N, \mathcal{F}_{\mathrm{BE}})}$ and combining all three bounds yields:
\begin{equation}
    \|Q_K - Q^\Pi\|_{2, d^\Pi} \leq 2\gamma^K \frac{R_{\max}}{1-\gamma} + \frac{1}{1 - \sqrt{\gamma}} \sqrt{C_{\mathrm{state}} \cdot C \cdot \epsilon_{\mathrm{stat}}(N, \mathcal{F}_{\mathrm{BE}})} + \frac{2 \epsilon_p R_{\max}}{(1 - \epsilon_p)(1 - \gamma)^2}.
\end{equation}
The $\lesssim$ form follows from the algebraic identity $\frac{1}{1-\sqrt{\gamma}} = \frac{1+\sqrt{\gamma}}{1-\gamma} \leq \frac{2}{1-\gamma}$ and absorbing the constants $2$ and $1/(1-\epsilon_p)$.
\end{proof}

\subsubsection{IO-FQE-AC}

\begin{assumption}[Finite-sample conditions for IO-FQE-AC]
\label{assump:finite_ac}
In addition to Assumption~\ref{assump:coverage_ac}: (a) the function class $\mathcal{F}_{\mathrm{AC}} \subset \{Q: \sS \times \sO \times \sA \to \mathbb{R}\}$ satisfies realizability ($Q^\Pi \in \mathcal{F}_{\mathrm{AC}}$) and Bellman completeness ($\mathcal{T}_{\mathrm{AC}} Q \in \mathcal{F}_{\mathrm{AC}}$ for all $Q \in \mathcal{F}_{\mathrm{AC}}$); (b) the Monte Carlo marginalization uses $M$ samples per option per transition.
\end{assumption}

Define the intra-option concentrability coefficient
\begin{equation}
    \label{eq:concentrability_io}
    C_{\mathrm{IO}} \triangleq \sup_{s, o, a} \frac{d^\Pi(s, o, a)}{d_\mathcal{D}(s)\, p_\mathcal{D}(a \mid s)},
\end{equation}
where $d^\Pi(s, o, a) = d^\Pi(s, o)\, \pi_o(a \mid s)$ is the normalized discounted state-option-action visitation distribution induced by $\Pi$, and $d_\mathcal{D}(s)\, p_\mathcal{D}(a \mid s)$ is the state-action marginal of the data distribution. This coefficient measures the distributional shift between the state-action-option occupancy of $\Pi$ and the data distribution.

\begin{theorem}[Error Bound for IO-FQE-AC]
\label{thm:error_ac}
Under Assumptions~\ref{assump:coverage_ac} and~\ref{assump:finite_ac}, suppose that each unweighted regression step satisfies
\begin{equation}
    \|Q_k - \hat{\mathcal{T}}_{\mathrm{AC}} Q_{k-1}\|_{2, d_\mathcal{D} p_\mathcal{D}} \leq \sqrt{\epsilon_{\mathrm{stat}}(N, \mathcal{F}_{\mathrm{AC}})}.
\end{equation}
Then, in expectation over the $M$ Monte Carlo samples drawn at each bootstrapping step, the output $Q_K$ after $K$ iterations satisfies
\begin{equation}
    \mathbb{E}_{\mathrm{MC}}\big[\|Q_K - Q^\Pi\|_{2, d^\Pi_{\mathrm{AC}}}\big] \leq 2\gamma^K \frac{R_{\max}}{1 - \gamma} + \frac{1}{1 - \sqrt{\gamma}} \sqrt{C_{\mathrm{IO}} \cdot \epsilon_{\mathrm{stat}}(N, \mathcal{F}_{\mathrm{AC}})} + \frac{1}{1 - \sqrt{\gamma}} \frac{\gamma R_{\max}}{(1 - \gamma)\sqrt{M}}.
\end{equation}
Consequently, the expected error bound simplifies to:
\begin{equation}
    \mathbb{E}_{\mathrm{MC}}\big[\|Q_K - Q^\Pi\|_{2, d^\Pi_{\mathrm{AC}}}\big] \lesssim \gamma^K \frac{R_{\max}}{1 - \gamma} + \frac{1}{1 - \gamma} \underbrace{\sqrt{C_{\mathrm{IO}} \cdot \epsilon_{\mathrm{stat}}(N, \mathcal{F}_{\mathrm{AC}})}}_{\text{extrapolated regression error}} + \frac{1}{(1 - \gamma)^2} \underbrace{\frac{R_{\max}}{\sqrt{M}}}_{\text{MC marginalization error}}.
\end{equation}
\end{theorem}

\begin{proof}
The proof follows the same unrolling and change-of-measure framework as Theorem~\ref{thm:error_be}, adapted to the augmented state-option-action space $\sS \times \sO \times \sA$.

Let $Q_k(s, o, a)$ denote the action-conditioned value function at iteration $k$, and $\mathcal{T}_{\mathrm{AC}}$ the exact population operator defined in Proposition~\ref{prop:bellman_consistency_ac}. The algorithm uses the empirical operator $\hat{\mathcal{T}}_{\mathrm{AC}}$ where the exact marginalization $\mathbb{E}_{a \sim \pi_o}$ is replaced by an $M$-sample Monte Carlo estimate.

\textbf{Step 1: Per-iteration error decomposition.}
Define the point-wise Bellman residual $e_k(s, o, a) \triangleq Q_k(s, o, a) - (\mathcal{T}_{\mathrm{AC}} Q_{k-1})(s, o, a)$. We decompose this into the statistical regression error $\epsilon_k$ and the MC marginalization error $\Delta_M$:
\begin{equation}
    e_k(s, o, a) = \underbrace{Q_k(s, o, a) - (\hat{\mathcal{T}}_{\mathrm{AC}} Q_{k-1})(s, o, a)}_{\triangleq \, \epsilon_k(s, o, a)} \;+\; \underbrace{(\hat{\mathcal{T}}_{\mathrm{AC}} Q_{k-1})(s, o, a) - (\mathcal{T}_{\mathrm{AC}} Q_{k-1})(s, o, a)}_{\triangleq \, \Delta_M(s, o, a)}.
\end{equation}
By assumption, the regression error satisfies $\|\epsilon_k\|_{2, d_\mathcal{D} p_\mathcal{D}} \leq \sqrt{\epsilon_{\mathrm{stat}}(N, \mathcal{F}_{\mathrm{AC}})}$.

The term $\Delta_M$ denotes the MC perturbation at iteration $k$ (suppressing the index $k$ for brevity); each perturbation satisfies the same bound because $Q_{k-1}$ is uniformly bounded by $R_{\max}/(1-\gamma)$. The MC error arises from the target computation: $\hat{y}_o - y_o = \gamma\,\mathbb{E}_{s'}[\hat{U}_{Q_{k-1}}(s', o) - U_{Q_{k-1}}(s', o)]$, where $\hat{U}$ uses the $M$-sample Monte Carlo estimate $\hat{Q}_{k-1}(s', o) = \frac{1}{M}\sum_{m=1}^M Q_{k-1}(s', o, a'_m)$ in place of the exact marginalization $Q_{k-1}(s', o) = \mathbb{E}_{\pi_o}[Q_{k-1}(s', o, a')]$. Since $|Q_{k-1}(s', o, a')| \leq R_{\max}/(1-\gamma)$, the variance of the sample mean is bounded by $R_{\max}^2/((1-\gamma)^2 M)$, and the $\gamma$ factor from the target gives, for any distribution including $d^\Pi_{\mathrm{AC}}(s, o, a) \triangleq d^\Pi(s, o)\pi_o(a \mid s)$:
\begin{equation}
    \mathbb{E}_{\mathrm{MC}}\big[\|\Delta_M\|_{2, d^\Pi_{\mathrm{AC}}}^2\big] \leq \frac{\gamma^2 R_{\max}^2}{(1-\gamma)^2 M}.
\end{equation}

\textbf{Step 2: Operator unrolling.}
Define the exact Markov transition kernel $P^\Pi_{\mathrm{AC}}$ on $\sS \times \sO \times \sA$ such that $(P^\Pi_{\mathrm{AC}} f)(s, o, a) \triangleq \mathbb{E}_{s' \sim P(\cdot \mid s, a)}[U_f(s', o)]$, where $U_f$ marginalizes the next action over $\pi_{o'}$. 
Unrolling $K$ iterations and taking the absolute value pointwise yields:
\begin{equation}
    |Q_K - Q^\Pi| \leq \sum_{k=1}^K \gamma^{K-k} (P^\Pi_{\mathrm{AC}})^{K-k} |\epsilon_k| \;+\; \sum_{k=1}^K \gamma^{K-k} (P^\Pi_{\mathrm{AC}})^{K-k} |\Delta_M| \;+\; \gamma^K (P^\Pi_{\mathrm{AC}})^K |Q_0 - Q^\Pi|.
\end{equation}

\textbf{Step 3: Norm conversion and change of measure.}
Taking the $L_2(d^\Pi_{\mathrm{AC}})$ norm, we bound the three components using Jensen's inequality $[(P^\Pi_{\mathrm{AC}})^m |f|]^2 \leq (P^\Pi_{\mathrm{AC}})^m (f^2)$. By the discounted occupancy property, $d^\Pi_{\mathrm{AC}} (P^\Pi_{\mathrm{AC}})^m \leq \gamma^{-m} d^\Pi_{\mathrm{AC}}$.

For the initial error term, $\|Q_0 - Q^\Pi\|_\infty \leq 2R_{\max}/(1-\gamma)$, yielding a contribution of $2\gamma^K R_{\max}/(1-\gamma)$.

For the regression error $\epsilon_k$, we apply the change of measure using the intra-option concentrability coefficient $C_{\mathrm{IO}} = \sup \frac{d^\Pi_{\mathrm{AC}}}{d_\mathcal{D} p_\mathcal{D}}$:
\begin{align}
    \big\| (P^\Pi_{\mathrm{AC}})^m |\epsilon_k| \big\|_{2, d^\Pi_{\mathrm{AC}}}^2 &\leq \gamma^{-m} \int d^\Pi_{\mathrm{AC}} \epsilon_k^2 \nonumber \\
    &\leq \gamma^{-m} C_{\mathrm{IO}} \int d_\mathcal{D} p_\mathcal{D} \epsilon_k^2 \;=\; \gamma^{-m} C_{\mathrm{IO}} \|\epsilon_k\|_{2, d_\mathcal{D} p_\mathcal{D}}^2.
\end{align}
Taking the square root and summing over $k$:
\begin{equation}
    \sum_{k=1}^K \gamma^{K-k} \big\| (P^\Pi_{\mathrm{AC}})^{K-k} |\epsilon_k| \big\|_{2, d^\Pi_{\mathrm{AC}}} \leq \frac{1}{1 - \sqrt{\gamma}} \sqrt{C_{\mathrm{IO}} \cdot \epsilon_{\mathrm{stat}}(N, \mathcal{F}_{\mathrm{AC}})}.
\end{equation}

For the MC marginalization error $\Delta_M$, because its variance bound is established directly under $d^\Pi_{\mathrm{AC}}$ without involving the behavior data distribution, no concentrability shift is needed. Applying the occupancy bound and taking expectations:
\begin{align}
    \mathbb{E}_{\mathrm{MC}}\Big[ \big\| (P^\Pi_{\mathrm{AC}})^m |\Delta_M| \big\|_{2, d^\Pi_{\mathrm{AC}}}^2 \Big] &\leq \gamma^{-m} \mathbb{E}_{\mathrm{MC}}\big[\|\Delta_M\|_{2, d^\Pi_{\mathrm{AC}}}^2\big] \leq \gamma^{-m} \frac{\gamma^2 R_{\max}^2}{(1-\gamma)^2 M}.
\end{align}
By Jensen's inequality ($\mathbb{E}[X] \leq \sqrt{\mathbb{E}[X^2]}$), the expected norm is bounded by $\gamma^{-m/2} \cdot \frac{\gamma R_{\max}}{(1-\gamma)\sqrt{M}}$. Summing the geometric series:
\begin{equation}
    \mathbb{E}_{\mathrm{MC}} \Big[ \sum_{k=1}^K \gamma^{K-k} \big\| (P^\Pi_{\mathrm{AC}})^{K-k} |\Delta_M| \big\|_{2, d^\Pi_{\mathrm{AC}}} \Big] \leq \frac{1}{1 - \sqrt{\gamma}} \frac{\gamma R_{\max}}{(1-\gamma)\sqrt{M}}.
\end{equation}

Combining the three terms yields:
\begin{equation}
    \mathbb{E}_{\mathrm{MC}}\big[\|Q_K - Q^\Pi\|_{2, d^\Pi_{\mathrm{AC}}}\big] \leq 2\gamma^K \frac{R_{\max}}{1 - \gamma} + \frac{1}{1 - \sqrt{\gamma}} \sqrt{C_{\mathrm{IO}} \cdot \epsilon_{\mathrm{stat}}(N, \mathcal{F}_{\mathrm{AC}})} + \frac{1}{1 - \sqrt{\gamma}} \frac{\gamma R_{\max}}{(1 - \gamma)\sqrt{M}}.
\end{equation}
The $\lesssim$ form follows from $\frac{1}{1-\sqrt{\gamma}} = \frac{1+\sqrt{\gamma}}{1-\gamma} \leq \frac{2}{1-\gamma}$ and absorbing $\gamma \leq 1$.
\end{proof}

\paragraph{Comparison of the two bounds.}
Relative to the standard flat FQE error bound, both IO-FQE bounds contain an additional additive error term that reflects the cost of operating in the augmented space. In IO-FQE-BE, the regression error is amplified by $\sqrt{C_{\mathrm{state}} \cdot C}$, factoring the distributional shift into a state-level component ($C_{\mathrm{state}}$) and an action-level importance weighting component ($C$). The density estimation bias $\epsilon_p$ is an irreducible additive term, but no action extrapolation occurs during bootstrapping. In IO-FQE-AC, the regression error is amplified by $\sqrt{C_{\mathrm{IO}}}$, a single coefficient that jointly captures state and action distributional shift. The MC marginalization error is an independent additive term controlled by $M$.
\section{Experimental Details}
\label{app:experimental_details}

This section provides full implementation details for the experiments in Section~\ref{sec:experiments}. Section~\ref{app:env_options} describes the environments and option construction, Section~\ref{app:datasets} describes the dataset construction for both regimes, Section~\ref{app:ground_truth} reports the ground-truth policy values, Section~\ref{app:hyperparameters} lists the hyperparameter configuration, and Section~\ref{app:significance} reports the statistical significance of the main results.

\subsection{Environment and Option Construction}
\label{app:env_options}

\subsubsection{AntMaze}

The AntMaze domain uses a quadruped Ant agent with observation dimension $d=29$ and action dimension $m=8$. Three maze layouts are used: UMaze from D4RL~\cite{fu2020d4rl} and Medium and Large from OGBench~\cite{park2024ogbench}. In each layout, the agent navigates from a fixed start cell to a goal cell. The reward is $-1$ at each non-terminal step and $0$ upon reaching the goal, so the value function reflects the discounted steps-to-goal. 

\begin{figure}[t]
\centering
\begin{minipage}[b]{0.26\textwidth}
\centering
\includegraphics[width=\linewidth]{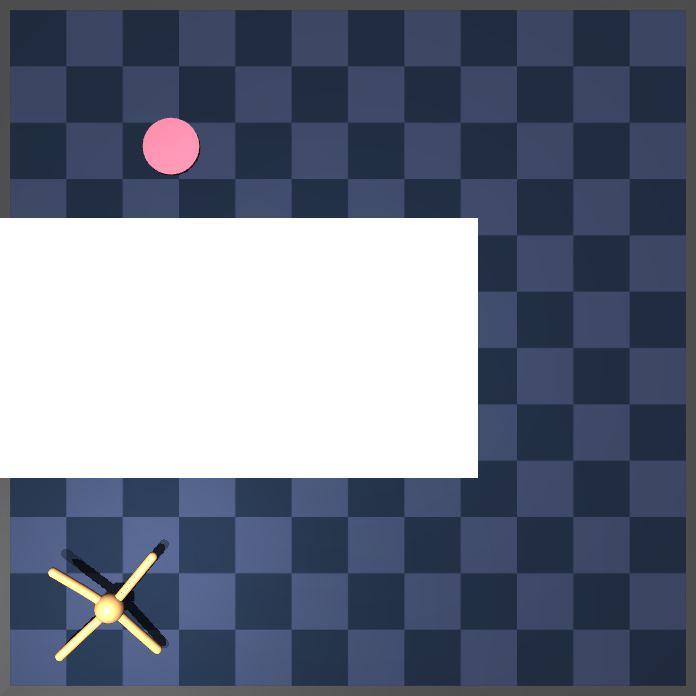}\\[2pt]
\small UMaze
\end{minipage}\hfill
\begin{minipage}[b]{0.26\textwidth}
\centering
\includegraphics[width=\linewidth]{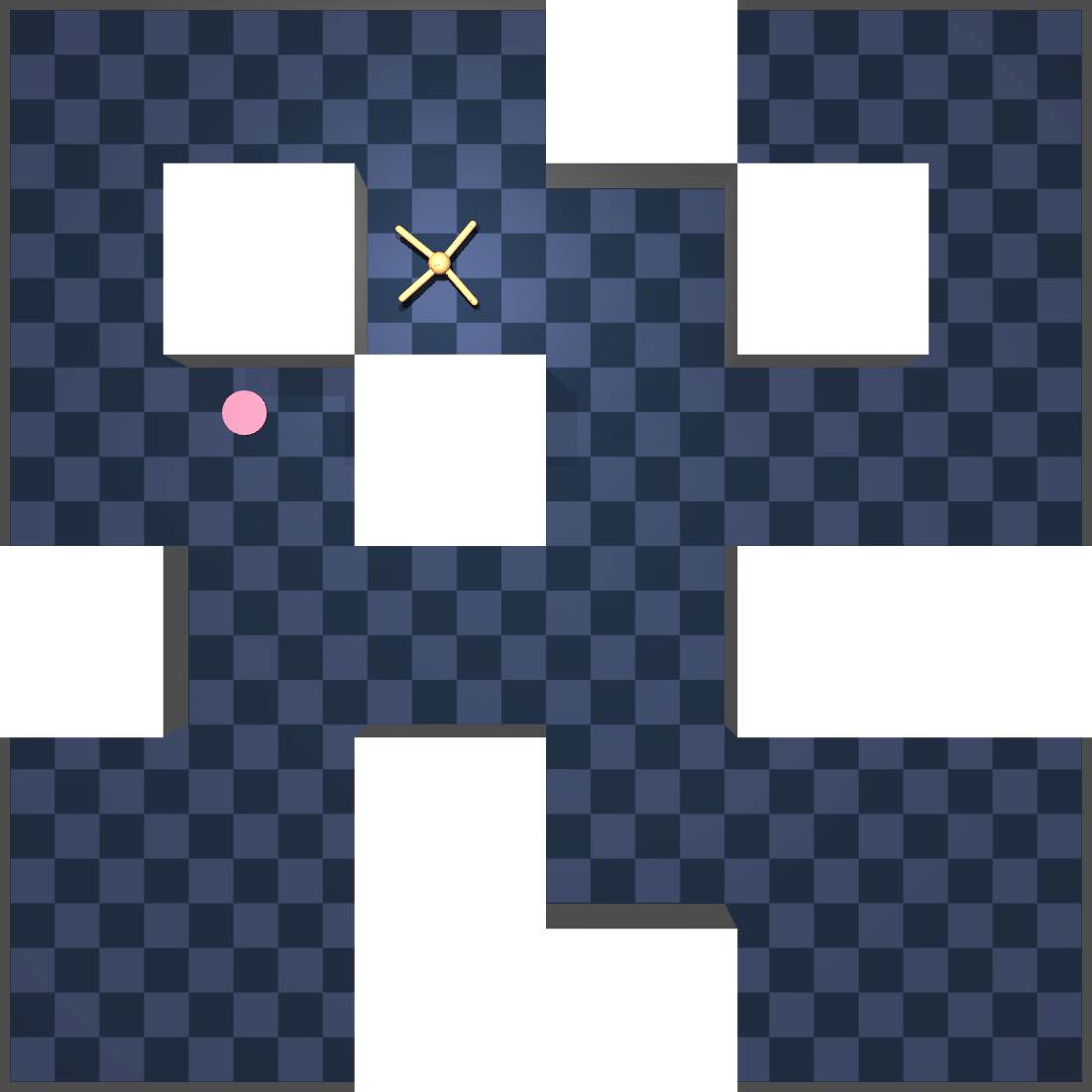}\\[2pt]
\small Medium
\end{minipage}\hfill
\begin{minipage}[b]{0.37\textwidth}
\centering
\includegraphics[width=\linewidth]{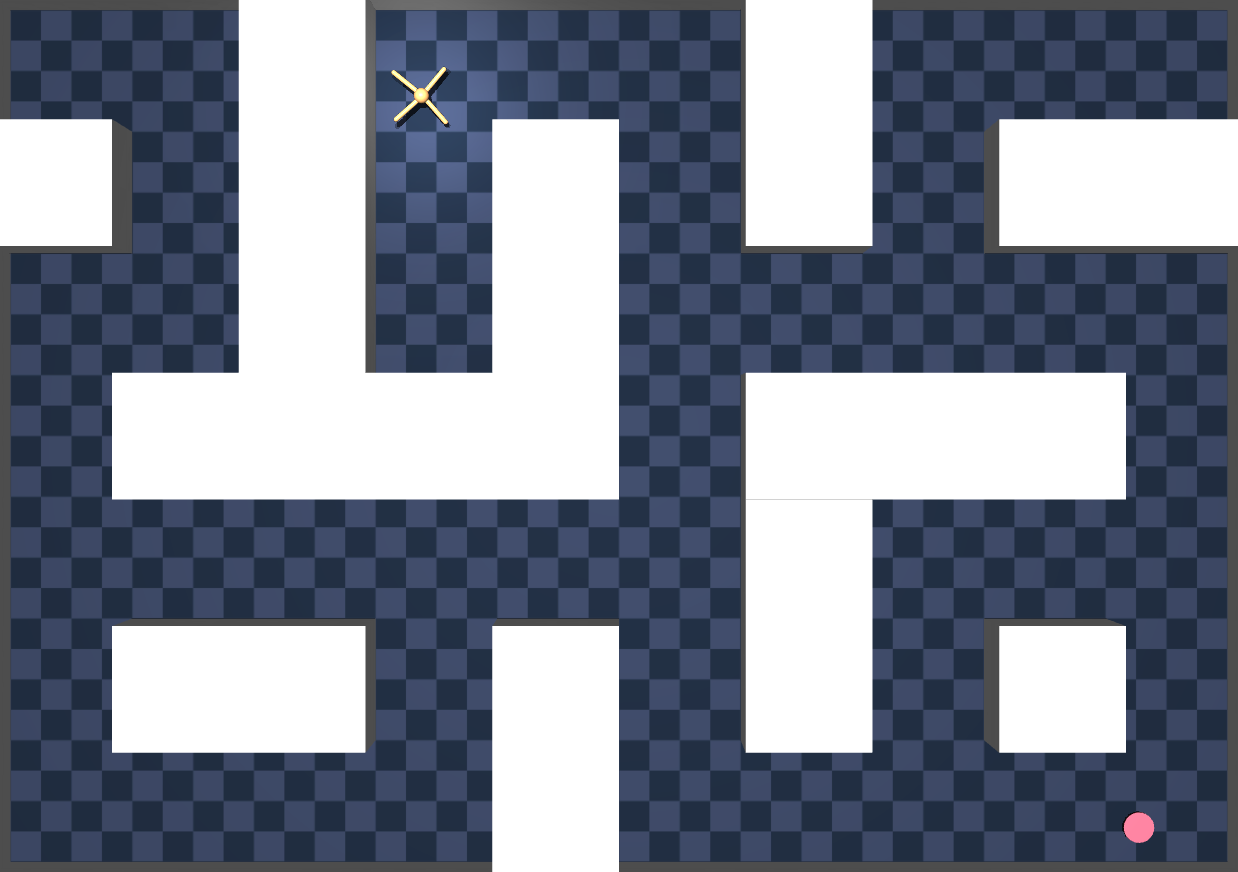}\\[2pt]
\small Large
\end{minipage}
\caption{AntMaze layouts used in our experiments: UMaze from D4RL~\cite{fu2020d4rl}, Medium and Large from OGBench~\cite{park2024ogbench}. The yellow figure is the Ant agent at its start position; the pink marker indicates the goal. Layouts are shown at their relative scales.}
\label{fig:antmaze-envs}
\end{figure}

\paragraph{Option set.}
The option set $\sO = \{o_\text{up}, o_\text{down}, o_\text{left}, o_\text{right}\}$ consists of four directional options corresponding to movement toward the adjacent waypoint in each cardinal direction. Each option $o_d$ is executed by a direction-conditioned locomotion controller trained with SAC~\cite{haarnoja2018soft}, which commands the agent to move at a fixed velocity in direction $d$. The option terminates deterministically when the agent's position comes within a fixed radius of the target waypoint ($\beta_{o_d}(s) = 1$ inside this radius, $\beta_{o_d}(s) = 0$ otherwise), a state condition consistent with the termination function of Section~\ref{sec:problem_formulation}. Data collection, ground-truth rollouts, and IO-FQE share this termination logic. 

\paragraph{High-level policy.}
The high-level policy $\mu(o \mid s)$ is trained by DQN~\cite{mnih2015human} over the waypoint graph of each maze layout. At each option-decision point, $\mu$ selects the directional option that moves the agent toward the next waypoint along the planned path.

\paragraph{Target policy.}
The target policy $\Pi = (\mu, \sO)$ uses an earlier checkpoint of the locomotion controllers (at 80\% of training) rather than the fully converged controllers, so that $\pi_o^{\text{target}} \neq \pi_o^{\text{behavior}}$ for each option $o$. The high-level policy $\mu$ is shared between target and behavior.

\subsubsection{Puzzle}

The Puzzle domain uses the OGBench Lights Out environment~\cite{park2024ogbench}, in which a robot arm manipulates buttons on a grid. Pressing a button toggles the state of that button and its neighbors. The goal is to reach a target button configuration. The observation space is $\mathbb{R}^{55}$ for $3\!\times\!3$ and $\mathbb{R}^{115}$ for $4\!\times\!6$, and the action space is $\mathbb{R}^5$ (end-effector delta position and gripper command). Actions are fed directly into OGBench's operational-space controller with no inverse kinematics. Two grid sizes are used: $3\!\times\!3$ (9 buttons) and $4\!\times\!6$ (24 buttons). The episode terminates successfully when all buttons match the goal configuration.

\begin{figure}[t]
\centering
\begin{minipage}[b]{0.35\textwidth}
\centering
\includegraphics[width=\linewidth]{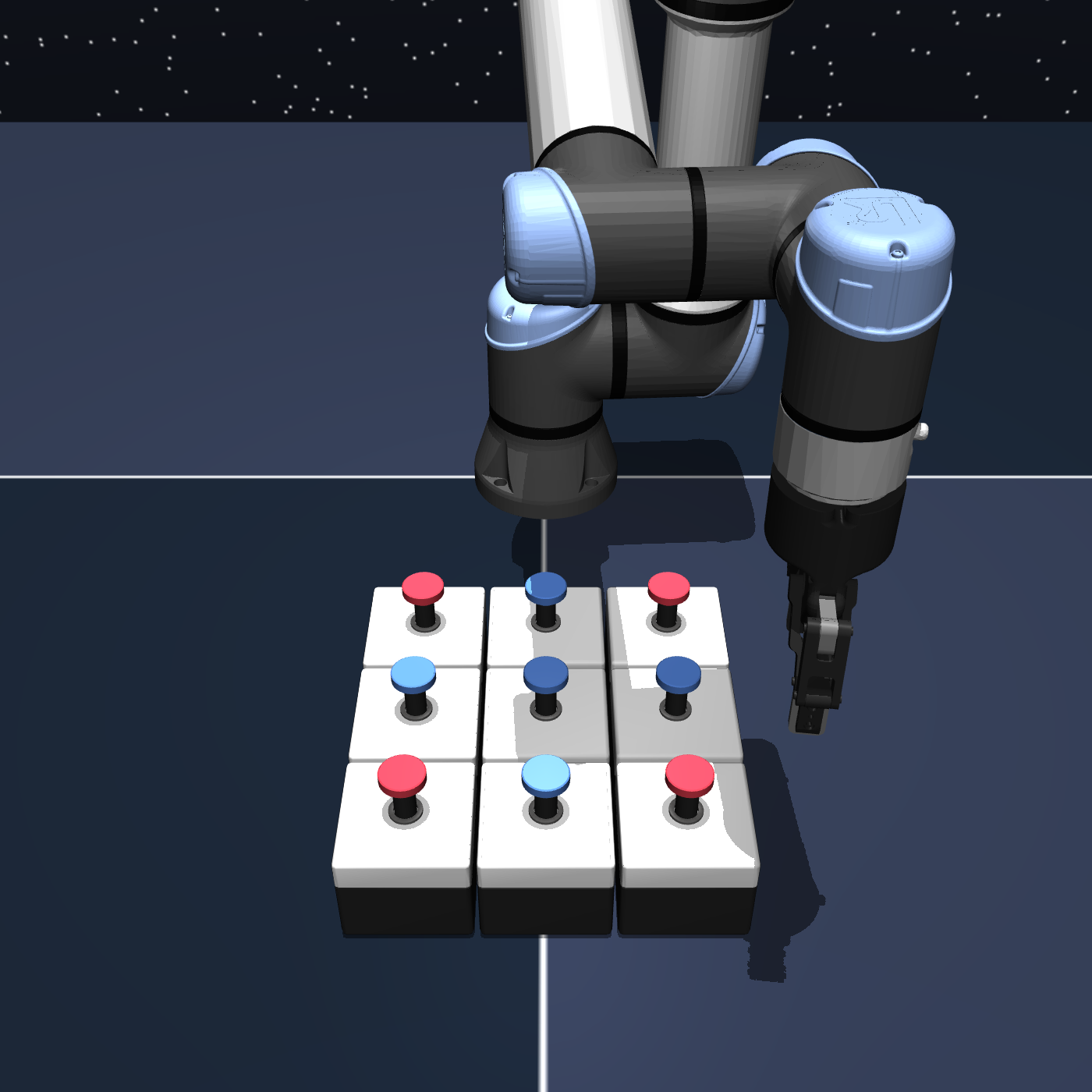}\\[2pt]
\small puzzle-$3\!\times\!3$
\end{minipage}\hfill
\begin{minipage}[b]{0.35\textwidth}
\centering
\includegraphics[width=\linewidth]{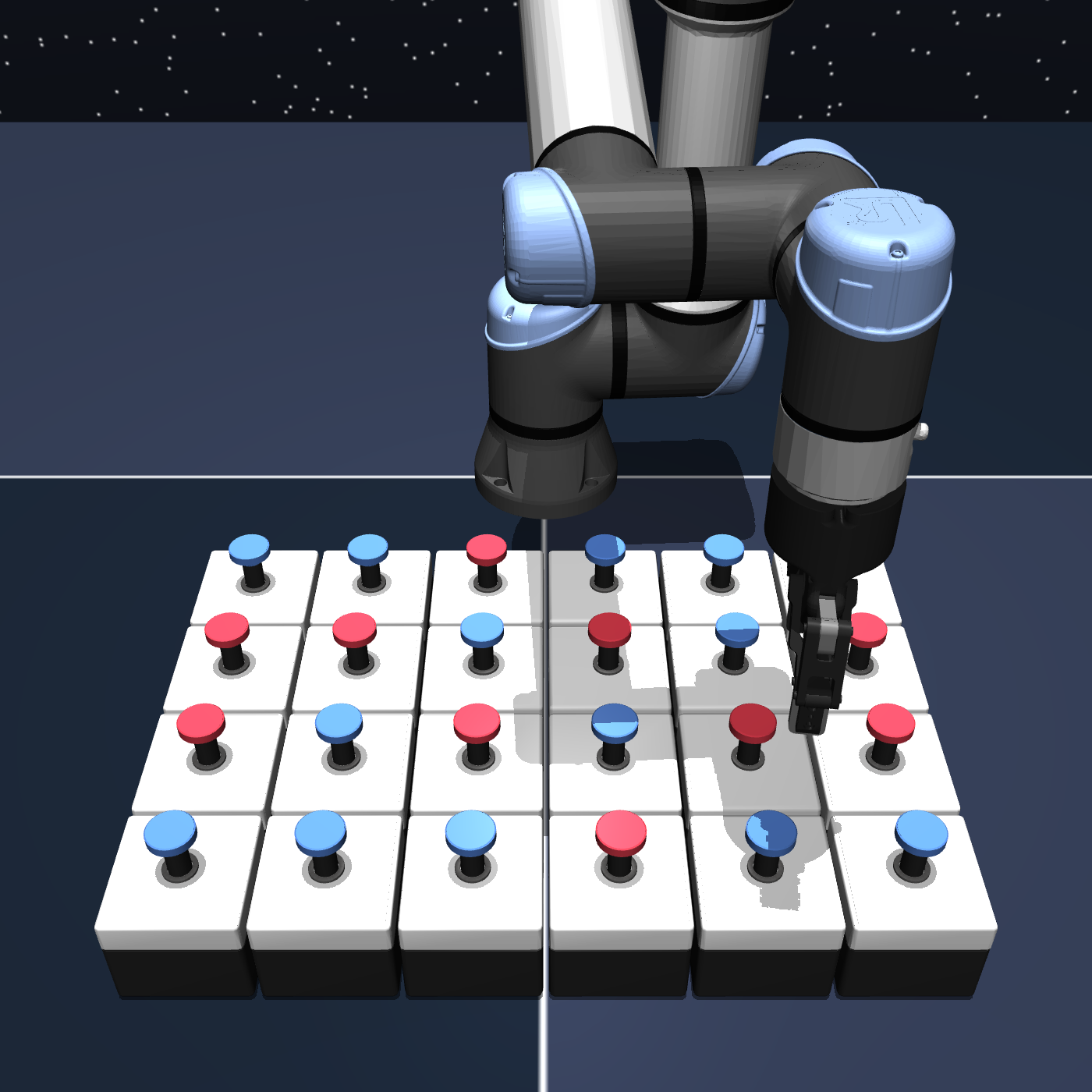}\\[2pt]
\small puzzle-$4\!\times\!6$
\end{minipage}
\caption{Puzzle (Lights Out) environments from OGBench~\cite{park2024ogbench}: $3\!\times\!3$ (9 buttons) and $4\!\times\!6$ (24 buttons). Pressing a button toggles its own state and that of its 4-connected neighbors. The button colors shown are the initial states under the rendering seed and do not carry semantic meaning.}
\label{fig:puzzle-envs}
\end{figure}

\paragraph{Option set.}
The option set $\sO = \{\text{Move}, \text{Press}, \text{Lift}\}$ consists of three manipulation primitives:
\begin{itemize}
    \item Move$(s, \text{target\_xy})$: a goal-conditioned policy (input dimension $55 + 2 = 57$) that moves the end-effector to a position above the target button.
    \item Press$(s)$: an unconditional policy (input dimension $55$) that executes a downward motion with gripper close.
    \item Lift$(s)$: an unconditional policy (input dimension $55$) that executes an upward motion with gripper open.
\end{itemize}
All three policies are trained via behavioral cloning on data segmented from rollouts of the OGBench \texttt{ButtonMarkovOracle} in \texttt{puzzle-3x3-v0} data-collection mode. The oracle's internal phase boundaries (end-effector within $0.04$\,m of the target button, button state flip, end-effector height exceeding $0.16$\,m) define the segmentation points, and the corresponding $(s, a)$ pairs from each phase are used to train the respective skill policy.

\paragraph{Termination conditions.}
Each option terminates deterministically upon satisfaction of a geometric or state condition:
\begin{itemize}
    \item Move: $\| \text{EE}_{xy} - \text{button}_{xy} \| \leq 0.04$\,m (end-effector reaches target button).
    \item Press: the target button state matches the goal state and $\| \text{EE}_{xy} - \text{target}_{xy} \| \leq 0.04$\,m (button toggled while end-effector is positioned above it).
    \item Lift: $\text{EE}_z > 0.16$\,m (end-effector raised to safe height).
\end{itemize}
These thresholds are consistent with the phase boundaries of the \texttt{ButtonMarkovOracle}.

\paragraph{High-level policy.}
The high-level policy $\mu$ is a rule-based state machine that does not require training. At the start of each episode, a Lights Out solver (Gaussian elimination over $\text{GF}(2)$) computes the sequence of buttons to press. The state machine cycles through $\text{MOVE} \to \text{PRESS} \to \text{LIFT}$ for each button in the sequence, advancing to the next button upon completion of each cycle. With probability $\varepsilon = 0.05$, $\mu$ selects uniformly among all available options instead of following the planned sequence.

\paragraph{Target policy.}
The target hierarchical policy uses a different set of intra-option policies (trained from a different behavioral cloning run) than the behavior hierarchical policy, so that $\pi_o^{\text{target}} \neq \pi_o^{\text{behavior}}$ for each option $o$. The high-level state machine structure is shared between target and behavior. 

\subsection{Dataset Construction}
\label{app:datasets}

\subsubsection{Flat Regime}

In the flat regime, the behavior policy is a flat primitive policy with no option structure. For AntMaze, this policy is obtained by training IQL~\cite{kostrikov2021offline} on the native D4RL or OGBench offline dataset for each maze layout, selecting whichever achieves higher return. For Puzzle, the behavior policy is obtained by training IQL on the OGBench noisy dataset for each grid size. In both cases, the resulting policy is Markovian by construction, and the collected data contain no option commitment structure.

\subsubsection{Hierarchical Regime}

In the hierarchical regime, data are collected by executing a hierarchical behavior policy in each environment.

For AntMaze, the behavior policy uses the same option set $\sO$ as the target but with the fully trained locomotion controllers. The high-level behavior strategy selects options as follows: with probability $1 - \varepsilon$ ($\varepsilon = 0.3$), the policy follows the shortest path on the maze graph via breadth-first search and selects the directional option toward the next waypoint; with probability $\varepsilon$, it selects uniformly among directions leading to adjacent open cells.

For Puzzle, the behavior policy uses the same state machine structure as the target but with different intra-option policies (the fully trained BC skill controllers). 

In both domains, the behavior process maintains option commitment across the full option duration. During data collection, option identities and termination events are additionally recorded at each step to enable the application of Option-PDIS; IO-FQE and flat FQE do not use these annotations.

\subsubsection{Dataset Sizes}

Table~\ref{tab:dataset_sizes} reports the number of episodes and mean episode length in each dataset. All environments use 200 episodes per regime. Episode length varies with maze complexity and puzzle grid size.

\begin{table}[h]
\centering
\footnotesize
\caption{Dataset sizes by environment and regime. Each dataset consists of 200 episodes.}
\label{tab:dataset_sizes}
\begin{tabular}{@{}l cc cc@{}}
\toprule
& \multicolumn{2}{c}{Flat} & \multicolumn{2}{c}{Hierarchical} \\
\cmidrule(lr){2-3} \cmidrule(lr){4-5}
Environment & Episodes & Mean Length & Episodes & Mean Length \\
\midrule
AntMaze-U              & 200 & 412.6    & 200  & 404.3 \\
AntMaze-M              & 200 & 504.7    & 200  & 477.5 \\
AntMaze-L              & 200 & 764.7    & 200  & 733.2\\
Puzzle $3\!\times\!3$   & 200 & 4{,}579.1 & 200 & 4{,}764.3 \\
Puzzle $4\!\times\!6$   & 200 & 6{,}753.8 & 200 & 6{,}488.6 \\
\bottomrule
\end{tabular}
\end{table}

\subsection{Ground-Truth Policy Values}
\label{app:ground_truth}

Table~\ref{tab:ground_truth} reports the Monte Carlo estimates of $J(\Pi)$ and $J(\pi_M)$ used as evaluation targets throughout the experiments. Each value is computed from 2000 independent rollouts.

\begin{table}[t]
\centering
\footnotesize
\caption{Ground-truth policy values estimated by Monte Carlo rollout. Values report mean $\pm$ one standard deviation.}
\label{tab:ground_truth}
\begin{tabular}{@{}lcc@{}}
\toprule
Environment & $J(\Pi)$ & $J(\pi_M)$ \\
\midrule
AntMaze-U           & $-94.30 \pm 0.32$    & $-94.87 \pm 0.28$ \\
AntMaze-M           & $-95.24 \pm 0.28$    & $-98.35 \pm 0.57$ \\
AntMaze-L           & $-96.42 \pm 0.41$    & $-98.78 \pm 0.62$ \\
Puzzle $3 \times 3$ & $-226.52 \pm 3.90$   & $-285.35 \pm 4.53$ \\
Puzzle $4 \times 6$ & $-895.33 \pm 32.14$ & $-801.64 \pm 29.33$ \\
\bottomrule
\end{tabular}
\end{table}

\subsection{Hyperparameter Configuration}
\label{app:hyperparameters}

Table~\ref{tab:hyperparameters} lists the configuration shared by all FQE-based estimators (flat FQE, IO-FQE-BE, and IO-FQE-AC) on AntMaze and Puzzle. This configuration is fixed from standard defaults before any evaluation. The mixture density network of IO-FQE-BE is selected per environment by held-out likelihood alone ($K = 3$ components in the setting of Table~\ref{tab:ablation_density}), and its importance weights use cross-option self-normalization without clipping (Appendix~\ref{app:algorithms}). Option-PDIS with INCRIS has no manually tuned hyperparameters: importance ratios are computed exactly from the known behavior and target policy densities, and INCRIS selects its truncation internally by its estimated-MSE criterion~\citep[Alg.~1]{guo2017using}.

\begin{table}[h]
\centering
\footnotesize
\caption{Hyperparameter configuration shared by all FQE-based estimators on AntMaze and Puzzle, fixed from standard defaults before any evaluation.}
\label{tab:hyperparameters}
\begin{tabular}{@{}ll@{}}
\toprule
Hyperparameter & Value \\
\midrule
Discount factor $\gamma$ & $0.99$ \\
Network & MLP with hidden layers $[256, 256]$ \\
Optimizer & AdamW (learning rate $3 \times 10^{-4}$, weight decay $10^{-5}$) \\
Target-network update rate $\tau$ & $0.005$ \\
Gradient steps & $10^5$ \\
Batch size & $256$ \\
MC action samples $M$ (flat FQE, IO-FQE-AC) & $10$ \\
Behavior density model (IO-FQE-BE) & MDN, selected by held-out likelihood \\
Importance weights (IO-FQE-BE) & cross-option self-normalization, no clipping \\
\bottomrule
\end{tabular}
\end{table}

\subsection{Statistical Significance of the Main Results}
\label{app:significance}

All methods within a row of Table~\ref{tab:results} are evaluated on the same ten dataset seeds, so differences between methods are assessed with paired bootstrap comparisons that jointly resample the seed indices ($10^4$ resamples). Each comparison yields a 95\% confidence interval for the paired MSE difference. Two sets of comparisons are performed. First, the more accurate IO-FQE variant is compared with each applicable baseline evaluated against $J(\Pi)$ (Option-PDIS and flat FQE); the flat FQE column evaluated against $J(\pi_M)$ is excluded because it targets a different estimand. In every environment and regime, the more accurate IO-FQE variant significantly outperforms every applicable baseline: each confidence interval excludes zero. Second, the numerically best method is compared with the runner-up among all methods evaluated against $J(\Pi)$, including the other IO-FQE variant. These differences are not significant in every row, so Table~\ref{tab:results} marks no method in bold.
\section{Additional Ablation Study}

This section reports additional empirical studies that complement the main results in Section~\ref{sec:experiments}. Section~\ref{app:ablation_tabular} validates Corollary~\ref{cor:irreducibility} in a tabular setting where Bellman fixed points are computed exactly. Section~\ref{app:ablation_density} examines the sensitivity of IO-FQE-BE to behavior density estimation quality. Section~\ref{app:time_dependent_termination} evaluates IO-FQE on options whose termination depends on the elapsed time. Section~\ref{app:short_horizon} compares the estimators in a short-horizon setting with discrete actions. Section~\ref{app:mc_samples} examines the sensitivity of IO-FQE-AC to the number of Monte Carlo samples.

\subsection{Tabular Validation of Corollary~\ref{cor:irreducibility}}
\label{app:ablation_tabular}

The main experiments evaluate IO-FQE in deep offline RL settings, where flat FQE's error against $J(\Pi)$ combines the structural bias $\Delta(\Pi)$ with function approximation and finite-sample estimation effects. This ablation isolates the structural component by reproducing the comparison in a tabular setting, in which Bellman fixed points are computed exactly and the only remaining source of error is the targeted policy itself.

The environment is a $5\!\times\!5$ deterministic gridworld with two terminals: an upper-right cell with reward $+1$ and a lower-left cell with reward $0$, $\gamma = 0.9$, and a uniform initial distribution over non-terminal cells. The option set $\mathcal{O} = \{o_0, o_1\}$ contains two deterministic shortest-path options targeting the upper-right and lower-left terminals respectively, with strict commitment ($\beta_o(s) = 0$ except at terminals). On the interior cells the two options select opposing primitive actions, so any flat Markov projection must average over conflicting action distributions. The high-level policy is state-independent with $\mu(o_0 \mid s) = p$, and the flat projection is $\pi_M(a \mid s) = p \cdot \pi_{o_0}(a \mid s) + (1-p) \cdot \pi_{o_1}(a \mid s)$. Data are collected by a uniform random primitive behavior policy, satisfying Assumption~\ref{assump:flat_data}. Both $J(\Pi)$ and $J(\pi_M)$ are computed by exact Bellman fixed-point solves rather than Monte Carlo estimation.

\begin{wrapfigure}{r}{0.55\textwidth}
\vspace{-1.2em}
\centering
\includegraphics[width=0.55\textwidth]{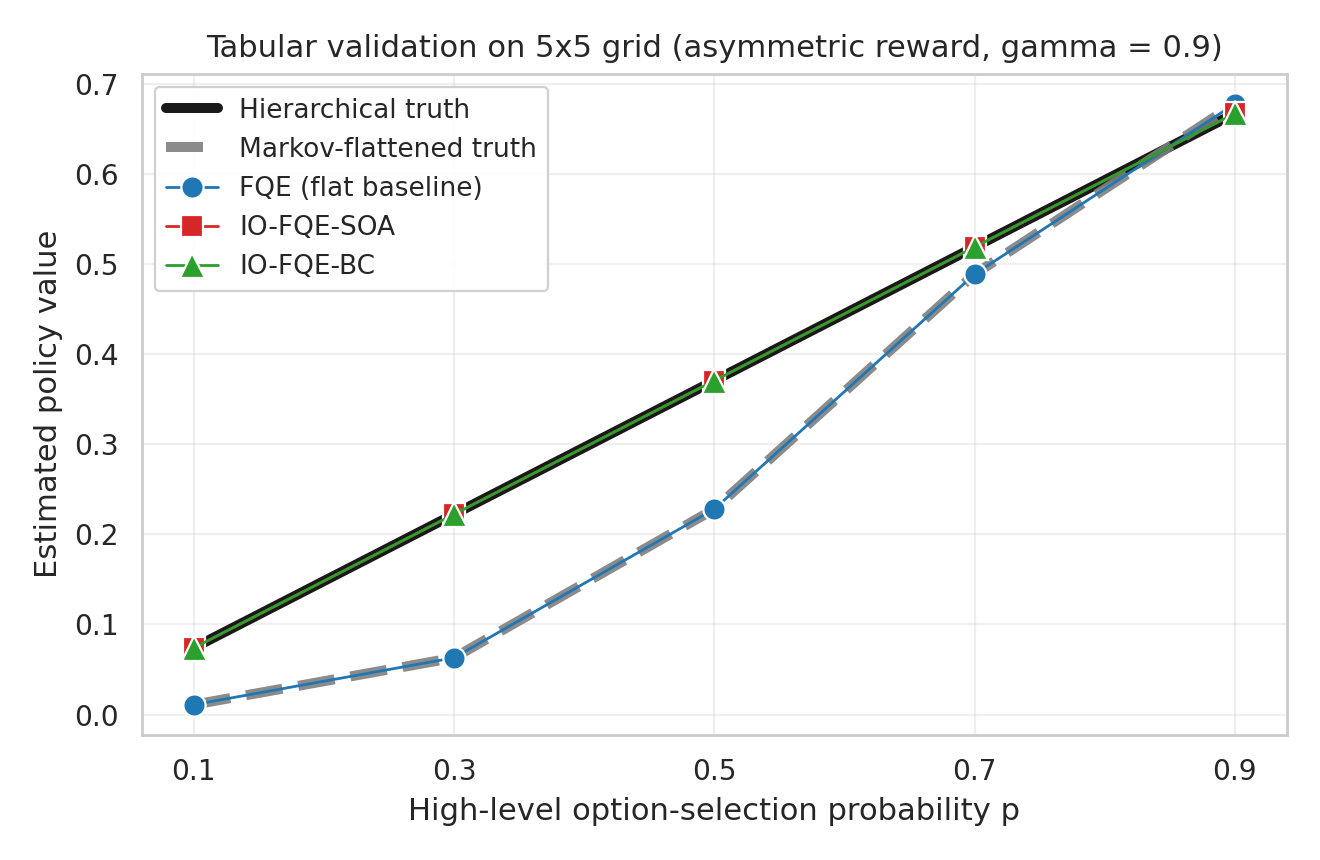}
\caption{Tabular validation on a $5\!\times\!5$ deterministic gridworld with two strict-commitment options. Left: estimated $J(\Pi)$ as $p = \mu(o_0 \mid s)$ varies; flat FQE tracks $J(\pi_M)$ while both IO-FQE variants recover $J(\Pi)$. Right: absolute bias vs.\ dataset size at $p = 0.5$; flat FQE's bias plateaus at $|\Delta(\Pi)|$ regardless of $N$. Mean over 5 seeds; standard deviation is below $10^{-11}$.}
\label{fig:tabular_validation}
\vspace{-1em}
\end{wrapfigure}

Figure~\ref{fig:tabular_validation} reports two sweeps. The left panel sweeps $p \in \{0.1, 0.3, 0.5, 0.7, 0.9\}$ with $N = 5000$ episodes. Flat FQE's estimate tracks $J(\pi_M)$ at every $p$, confirming that flat FQE accurately evaluates the wrong target. The induced bias against $J(\Pi)$ is non-monotonic and asymmetric about $p = 0.5$, with the largest deviation at $p = 0.3$ ($|\Delta(\Pi)| \approx 0.16$, a $72\%$ relative error). This asymmetry reflects the reward geometry: when $\pi_M$ replaces option commitment with random walks over conflicting actions, the resulting trajectories are biased toward the lower-reward terminal, and the magnitude of this effect depends on $p$.

The right panel sweeps the dataset size $N$ at fixed $p = 0.5$. Flat FQE's bias plateaus at $|\Delta(\Pi)| \approx 0.143$ across all $N$, confirming that the bias is invariant to dataset size. Both IO-FQE variants attain bias at numerical precision ($\sim 10^{-11}$) from the smallest dataset size onward; under deterministic dynamics with full coverage from a uniform behavior policy, the empirical Bellman operator coincides with the population operator at modest sample sizes, so IO-FQE's error floor is determined by numerical precision rather than statistical estimation. IO-FQE-BE and IO-FQE-AC produce identical estimates to numerical precision, consistent with the observation that the two variants share the same population fixed point and differ only through the function approximation and finite-sample effects exercised in the main experiments.

Option-PDIS is omitted: given option annotations it correctly evaluates $J(\Pi)$ by construction~\citep{guo2017using}, and its tabular behavior under uniform behavior coverage matches IO-FQE up to numerical precision, providing no additional validation of Corollary~\ref{cor:irreducibility}, which concerns the structural bias of flat methods. Its high-variance behavior in continuous action spaces is examined in Section~\ref{sec:experiments}.

\subsection{Effect of Behavior Density Estimation on IO-FQE-BE}
\label{app:ablation_density}

IO-FQE-BE requires an estimate of the behavior action density $p_\mathcal{D}(a \mid s)$ to compute the importance weights $w_o(s,a) = \pi_o(a \mid s) / p_\mathcal{D}(a \mid s)$. The finite-sample bound in Theorem~\ref{thm:error_be} shows that the weight perturbation error $\epsilon_p$ introduces an additive bias of order $\epsilon_p \cdot R_{\max} / (1 - \gamma)^2$ that persists regardless of dataset size. This ablation isolates the effect of density estimation quality on IO-FQE-BE by varying the density model while holding all other components fixed.

Three density estimators are compared on AntMaze Large under both the hierarchical and flat data regimes. Under the hierarchical regime, the behavior data are collected by a hierarchical policy with option commitment, so the marginal behavior action density $p_\mathcal{D}(a \mid s)$ at each state is a mixture of four intra-option policies. Under the flat regime, the behavior policy is a single IQL-trained primitive policy, producing a unimodal action distribution at each state. Comparing the two regimes on the same environment with the same density models isolates the multimodality of the behavior process as the factor governing density estimation difficulty. All other hyperparameters (network architecture, learning rate, number of FQE iterations, target network update schedule) are identical to the main experiments. Results report MSE against the Monte Carlo ground truth $J(\Pi)$, mean with 95\% bootstrapped confidence intervals over 10 seeds.

\begin{table}[h]
\centering
\footnotesize
\caption{Effect of behavior density estimation on IO-FQE-BE (AntMaze Large). Oracle uses the true behavior policy network to compute $p_\mathcal{D}(a \mid s)$; MDN and Gaussian are trained on the offline dataset. The oracle is omitted under flat data because the IQL behavior policy is unimodal and both trained density models already approximate it well. MSE is computed against $J(\Pi)$; mean with 95\% CI over 10 seeds.}
\label{tab:ablation_density}
\begin{tabular}{@{}lcc@{}}
\toprule
& \multicolumn{2}{c}{MSE ($\downarrow$)} \\
\cmidrule(l){2-3}
Density Estimator & Hierarchical & Flat \\
\midrule
Oracle (true $p_\mathcal{D}$) & $\mathbf{0.35 \pm 0.11}$ & --- \\
MDN ($K=3$)                   & $1.51 \pm 0.91$ & $\mathbf{0.32 \pm 0.11}$ \\
Gaussian                      & $22.7 \pm 4.0$ & $0.57 \pm 0.12$ \\
\bottomrule
\end{tabular}
\end{table}

With the true behavior density ($\epsilon_p = 0$), IO-FQE-BE achieves an MSE of $0.35$ under hierarchical data, four times lower than the MDN-based estimate ($1.51$) used in the main experiments. The difference between the oracle and the MDN ($1.16$ out of $1.51$ total MSE) indicates that density estimation error contributes the larger share of IO-FQE-BE's error in this setting, consistent with the additive bias term in Theorem~\ref{thm:error_be}. The Gaussian produces an MSE of $22.7$ under hierarchical data, fifteen times higher than the MDN: a unimodal model cannot represent the mixture of four intra-option policies, and the misspecified importance weights degrade IO-FQE-BE's estimate accordingly. Under flat data, the same Gaussian produces an MSE of $0.57$, close to the MDN ($0.32$), because the IQL behavior policy is unimodal and a Gaussian density model approximates it adequately. The contrast between the two regimes indicates that the Gaussian's failure under hierarchical data is attributable to the multimodal structure of the behavior action density rather than to the dimensionality of the action space or the capacity of the density model.

\subsection{Time-Dependent Termination}
\label{app:time_dependent_termination}

Section~\ref{sec:problem_formulation} handles a termination condition that depends on the time elapsed since option initiation by augmenting the state with the elapsed time $h$, which IO-FQE enumerates internally rather than reading from the data. This experiment verifies the extension in its purest case, in which termination depends on the elapsed time alone.

The environment is a $15\!\times\!15$ four-rooms gridworld with stochastic slip and a stay action. Five waypoint-navigation options (four doorway options and one goal-room option) run for exactly $H$ steps regardless of arrival. The environment and the intra-option policies remain Markov, and only the termination condition depends on the elapsed time. Because of the stochastic dynamics, the same state can be reached at different elapsed times with different termination outcomes, so the evaluation must track the elapsed time. The behavior data are collected by a hierarchical behavior policy whose intra-option policies differ from those of the target policy. Option annotations are recorded for Option-PDIS only, and IO-FQE and flat FQE use the primitive transitions alone, consistent with Assumption~\ref{assump:flat_data}. Ground-truth values are computed by exact solves: $J(\Pi) = 0.2407$ for $H = 4$, $J(\Pi) = 0.1359$ for $H = 8$, and $J(\pi_M) = 0.2897$. Table~\ref{tab:time_dependent_termination} reports the MSE against $J(\Pi)$, mean with 95\% bootstrap confidence intervals over 50 independently generated datasets.

\begin{table}[h]
\centering
\footnotesize
\caption{MSE against $J(\Pi)$ under time-dependent termination ($15\!\times\!15$ four-rooms gridworld, options of duration $H$). Each entry reports the mean with a 95\% bootstrap confidence interval over 50 independently generated datasets.}
\label{tab:time_dependent_termination}
\begin{tabular}{@{}lcc@{}}
\toprule
& \multicolumn{2}{c}{MSE ($\downarrow$)} \\
\cmidrule(l){2-3}
Method & $H = 4$ & $H = 8$ \\
\midrule
Flat FQE    & $2.40\ [2.32, 2.49] \times 10^{-3}$ & $2.37\ [2.34, 2.39] \times 10^{-2}$ \\
Option-PDIS & $6.14\ [4.05, 8.56] \times 10^{-3}$ & $1.06\ [0.71, 1.47] \times 10^{-3}$ \\
IO-FQE-BE   & $1.5\ [1.0, 2.1] \times 10^{-5}$ & $1.2\ [0.8, 1.7] \times 10^{-4}$ \\
IO-FQE-AC   & $2.5\ [1.7, 3.5] \times 10^{-5}$ & $1.5\ [1.0, 2.1] \times 10^{-4}$ \\
\bottomrule
\end{tabular}
\end{table}

Flat FQE's error is the commitment gap itself: its estimate ($0.2897$) does not change with $H$, and the gap vanishes at $H = 1$, for which $\Pi$ coincides with $\pi_M$.

\subsection{Short Horizons and Discrete Actions}
\label{app:short_horizon}

The main experiments use long-horizon continuous-control tasks, in which per-step action-density ratios compound across the trajectory (Section~\ref{subsec:results}). This experiment evaluates the estimators in a regime that favors importance sampling, with short horizons and discrete actions.

The environment is a four-rooms gridworld at two scales, $15\!\times\!15$ and $9\!\times\!9$, with five discrete actions. Five landmark-navigation options terminate on state conditions, $\beta_o(s) = \mathbf{1}[s = g_o]$ with $g_o$ the landmark of option $o$ (Definition~\ref{def:options}), and follow $\varepsilon$-soft shortest-path intra-option policies. The behavior and target policies differ only in the high-level policy, the favorable shared-intra-option-policy case characterized by \citet[Corollary~1]{guo2017using}, so each per-decision weight of Option-PDIS accumulates about eight bounded high-level option-probability ratios rather than primitive-action ratios. Option annotations are recorded for Option-PDIS only, and all estimators are finite-sample, unlike the exact solves of Appendix~\ref{app:ablation_tabular}. Table~\ref{tab:short_horizon} reports the MSE against $J(\Pi)$, and against $J(\pi_M)$ for flat FQE, mean with 95\% bootstrap confidence intervals over 10 seeds.

\begin{table}[h]
\centering
\footnotesize
\caption{MSE ($\times 10^{-5}$) in the short-horizon four-rooms gridworld with discrete actions. Each entry reports the mean with a 95\% bootstrap confidence interval over 10 seeds.}
\label{tab:short_horizon}
\begin{tabular}{@{}l ccc cc@{}}
\toprule
& & & & \multicolumn{2}{c}{Flat FQE} \\
\cmidrule(l){5-6}
Scale & Option-PDIS & IO-FQE-BE & IO-FQE-AC & vs $J(\Pi)$ & vs $J(\pi_M)$ \\
\midrule
$15\!\times\!15$ & $33\ [22, 47]$ & $11\ [6.8, 17]$ & $\mathbf{7.0}\ [4.4, 11]$ & $1840\ [1820, 1860]$ & $1.1\ [0.74, 1.6]$ \\
$9\!\times\!9$ & $6.7\ [3.6, 11]$ & $5.1\ [3.0, 7.5]$ & $\mathbf{3.0}\ [1.8, 4.5]$ & $1850\ [1830, 1870]$ & $0.56\ [0.36, 0.79]$ \\
\bottomrule
\end{tabular}
\end{table}

Table~\ref{tab:short_horizon} supports three observations. First, Option-PDIS becomes competitive at the $9\!\times\!9$ scale: its confidence interval overlaps those of both IO-FQE variants, although its mean MSE remains higher, whereas at the $15\!\times\!15$ scale the intervals of both IO-FQE variants lie below that of Option-PDIS. Option-PDIS receives strictly stronger information than IO-FQE, namely option labels that IO-FQE does not use, and is thus a strong choice when the effective horizon is short and annotated data are available, while IO-FQE remains competitive under the weaker primitive-transition data interface. Second, discrete actions remove the continuous-action approximations that most sharply separate the two IO-FQE variants: density estimation reduces to tabular counts, and the marginalization over $\pi_o$ reduces to an exact sum over the five actions. The errors of the two variants are of the same order, and the larger separation between them in Table~\ref{tab:results} is consistent with these additional continuous-action estimation burdens, while the lift to $\sS \times \sO$ that removes $\Delta(\Pi)$ is common to both variants. Third, flat FQE accurately estimates $J(\pi_M)$ rather than $J(\Pi)$: its MSE against $J(\pi_M)$ is of order $10^{-5}$, whereas its MSE against $J(\Pi)$ is close to $\Delta(\Pi)^2$ ($\Delta(\Pi) = 0.134$ at $15\!\times\!15$ and $0.136$ at $9\!\times\!9$), empirically illustrating the commitment gap characterized by Proposition~\ref{prop:irreducible_bias} and Corollary~\ref{cor:irreducibility}.

\subsection{Number of Monte Carlo Samples in IO-FQE-AC}
\label{app:mc_samples}

IO-FQE-AC approximates the marginalization over $\pi_o$ with $M$ Monte Carlo samples (Section~\ref{subsec:iofqe_ac}). The resulting marginalization error is controlled by computation, but it is only one component of the error of IO-FQE-AC. This ablation varies $M \in \{1, 10, 50\}$ for IO-FQE-AC on AntMaze-L under the flat regime, with all other components fixed (Appendix~\ref{app:hyperparameters}). Table~\ref{tab:mc_samples} reports the MSE against $J(\Pi)$ under the evaluation protocol of Section~\ref{subsec:setup}, with IO-FQE-BE as a reference.

\begin{table}[h]
\centering
\footnotesize
\caption{MSE of IO-FQE-AC on AntMaze-L under the flat regime as the number of Monte Carlo samples $M$ varies, with IO-FQE-BE as a reference. Each entry reports the mean with a 95\% bootstrapped confidence interval.}
\label{tab:mc_samples}
\begin{tabular}{@{}lc@{}}
\toprule
Method & MSE ($\downarrow$) \\
\midrule
IO-FQE-AC ($M = 1$)  & $1.63\ [0.61, 2.87]$ \\
IO-FQE-AC ($M = 10$) & $0.89\ [0.39, 1.39]$ \\
IO-FQE-AC ($M = 50$) & $0.82\ [0.36, 1.28]$ \\
IO-FQE-BE            & $0.32\ [0.21, 0.43]$ \\
\bottomrule
\end{tabular}
\end{table}

Most of the observed improvement occurs between $M = 1$ and $M = 10$, with diminishing returns thereafter. Increasing $M$ therefore does not close the gap in mean MSE to IO-FQE-BE over the tested range, and the residual difference is consistent with the remaining regression, coverage, and extrapolation terms in the error of IO-FQE-AC. Under hierarchical data on AntMaze-L, IO-FQE-AC already attains a lower mean MSE than IO-FQE-BE at $M = 10$ (Table~\ref{tab:results}).

\clearpage

\newpage

\section*{NeurIPS Paper Checklist}

\begin{enumerate}

\item {\bf Claims}
    \item[] Question: Do the main claims made in the abstract and introduction accurately reflect the paper's contributions and scope?
    \item[] Answer: \answerYes{}
    \item[] Justification: The three contributions stated in the abstract and Section~\ref{sec:problem_formulation} (characterizing the option commitment of flat FQE in Proposition~\ref{prop:irreducible_bias}, deriving IO-FQE in the augmented state-option space, and developing the IO-FQE-BE and IO-FQE-AC variants) are formalized in Section~\ref{sec:problem_formulation}, derived in Section~\ref{sec:method}, and evaluated in Section~\ref{sec:experiments}.

\item {\bf Limitations}
    \item[] Question: Does the paper discuss the limitations of the work performed by the authors?
    \item[] Answer: \answerYes{}
    \item[] Justification: Limitations are discussed in Section~\ref{sec:discussion}. IO-FQE-BE depends on accurate behavior density estimation, which degrades as action dimensionality grows or behavior data concentrates along narrow trajectories; this dependence appears empirically in Section~\ref{sec:experiments} (the Puzzle-$4\!\times\!6$ hierarchical results and the data-efficiency curves) and theoretically as the density-estimation bias term in Theorem~\ref{thm:error_be}. IO-FQE-AC retains a Monte Carlo marginalization error controlled by the number of MC samples rather than the dataset size (Theorem~\ref{thm:error_ac}). Both variants further require every intra-option policy to be supported by the data action distribution at every visited state (Assumptions~\ref{assump:coverage_be} and~\ref{assump:coverage_ac}), and the per-transition computational cost scales linearly with $|\sO|$. The analysis additionally assumes the target's option structure---the option set, intra-option policies, and termination functions---is given rather than recovered from data, and the empirical evaluation covers two continuous-control simulation benchmarks (AntMaze and OGBench Puzzle).

\item {\bf Theory assumptions and proofs}
    \item[] Question: For each theoretical result, does the paper provide the full set of assumptions and a complete (and correct) proof?
    \item[] Answer: \answerYes{}
    \item[] Justification: All theoretical results are stated with their full assumption sets and accompanied by proofs. Proposition~\ref{prop:irreducible_bias}, which characterizes the option commitment gap of flat FQE, is stated in Section~\ref{sec:problem_formulation} under Assumption~\ref{assump:flat_data} with a complete proof in Appendix~\ref{app:proofs_sec2}. The Bellman consistency results for IO-FQE-BE and IO-FQE-AC (Proposition~\ref{prop:bellman_consistency_be} and Proposition~\ref{prop:bellman_consistency_ac}) are stated and proven in Appendix~\ref{app:bellman_consistency} under Assumption~\ref{assump:coverage_be} and Assumption~\ref{assump:coverage_ac}, respectively. The finite-sample error bounds in Theorem~\ref{thm:error_be} (IO-FQE-BE) and Theorem~\ref{thm:error_ac} (IO-FQE-AC) are stated in Appendix~\ref{app:error_bounds} under Assumptions~\ref{assump:coverage_be} and~\ref{assump:finite_be}, and Assumptions~\ref{assump:coverage_ac} and~\ref{assump:finite_ac}, respectively, with proof sketches that adapt the approximate policy iteration error propagation framework of \citet{munos2005error} to the augmented state-option space.

    \item {\bf Experimental result reproducibility}
    \item[] Question: Does the paper fully disclose all the information needed to reproduce the main experimental results of the paper to the extent that it affects the main claims and/or conclusions of the paper (regardless of whether the code and data are provided or not)?
    \item[] Answer: \answerYes{}
    \item[] Justification: Section~\ref{sec:experiments} describes the experimental setup, Appendix~\ref{app:algorithms} provides complete pseudocode for both IO-FQE variants, and Appendix~\ref{app:experimental_details} reports the environments, option construction, dataset construction, and method implementation details required to reproduce the main results.

\item {\bf Open access to data and code}
    \item[] Question: Does the paper provide open access to the data and code, with sufficient instructions to faithfully reproduce the main experimental results, as described in supplemental material?
    \item[] Answer: \answerNo{}
    \item[] Justification: To preserve double-blind anonymity, code is not released at submission time. The paper and appendix provide algorithmic pseudocode (Appendix~\ref{app:algorithms}) and full experimental specifications (Appendix~\ref{app:experimental_details}) to support reproduction; code will be released publicly upon acceptance.

\item {\bf Experimental setting/details}
    \item[] Question: Does the paper specify all the training and test details (e.g., data splits, hyperparameters, how they were chosen, type of optimizer) necessary to understand the results?
    \item[] Answer: \answerYes{}
    \item[] Justification: Section~\ref{sec:experiments} describes the experimental protocol, and Appendix~\ref{app:experimental_details} provides full details on environment construction, option construction, dataset construction, network architectures, optimizers, and hyperparameters for all methods evaluated.

\item {\bf Experiment statistical significance}
    \item[] Question: Does the paper report error bars suitably and correctly defined or other appropriate information about the statistical significance of the experiments?
    \item[] Answer: \answerYes{}
    \item[] Justification: All reported MSE values in Table~\ref{tab:results} and Figure~\ref{fig:data_efficiency} are accompanied by 95\% bootstrapped confidence intervals computed over 10 random seeds, with the variability capturing both random initialization and dataset subsampling.

\item {\bf Experiments compute resources}
    \item[] Question: For each experiment, does the paper provide sufficient information on the computer resources (type of compute workers, memory, time of execution) needed to reproduce the experiments?
    \item[] Answer: \answerYes{}
    \item[] Justification: All experiments were run on an internal academic cluster using single-GPU jobs, with each job allocated either an NVIDIA RTX~5090 or an NVIDIA A100 (40\,GB) depending on resource availability.
    
\item {\bf Code of ethics}
    \item[] Question: Does the research conducted in the paper conform, in every respect, with the NeurIPS Code of Ethics \url{https://neurips.cc/public/EthicsGuidelines}?
    \item[] Answer: \answerYes{}
    \item[] Justification: The research uses only publicly available simulation benchmarks (D4RL AntMaze and OGBench) and does not involve human subjects, sensitive personal data, or deployment to real-world systems.

\item {\bf Broader impacts}
    \item[] Question: Does the paper discuss both potential positive societal impacts and negative societal impacts of the work performed?
    \item[] Answer: \answerNA{}
    \item[] Justification: This paper develops foundational methodology for off-policy evaluation of hierarchical reinforcement learning policies. The contributions are theoretical and algorithmic, with no direct path to specific deployments that would carry foreseeable societal impacts.
    
\item {\bf Safeguards}
    \item[] Question: Does the paper describe safeguards that have been put in place for responsible release of data or models that have a high risk for misuse (e.g., pre-trained language models, image generators, or scraped datasets)?
    \item[] Answer: \answerNA{}
    \item[] Justification: The paper does not release pre-trained generative models, scraped datasets, or other assets that pose a risk for misuse.

\item {\bf Licenses for existing assets}
    \item[] Question: Are the creators or original owners of assets (e.g., code, data, models), used in the paper, properly credited and are the license and terms of use explicitly mentioned and properly respected?
    \item[] Answer: \answerYes{}
    \item[] Justification: All existing assets used in this work are publicly released and properly cited: D4RL~\citep{fu2020d4rl} (Apache License 2.0) for the AntMaze UMaze environment, OGBench~\citep{park2024ogbench} (MIT License) for the AntMaze Medium/Large and Puzzle environments and offline datasets, Stable-Baselines3~\citep{stable-baselines3} (MIT License) for the SAC implementation used to train locomotion controllers, and d3rlpy~\citep{Seno_d3rlpy_An_Offline_2022} (MIT License) for the IQL implementation used to train flat behavior policies. All assets are used in accordance with their licenses, which permit research use.

\item {\bf New assets}
    \item[] Question: Are new assets introduced in the paper well documented and is the documentation provided alongside the assets?
    \item[] Answer: \answerNA{}
    \item[] Justification: The paper does not release new datasets, models, or other assets at submission time.

\item {\bf Crowdsourcing and research with human subjects}
    \item[] Question: For crowdsourcing experiments and research with human subjects, does the paper include the full text of instructions given to participants and screenshots, if applicable, as well as details about compensation (if any)? 
    \item[] Answer: \answerNA{}
    \item[] Justification: The research does not involve crowdsourcing or human subjects.

\item {\bf Institutional review board (IRB) approvals or equivalent for research with human subjects}
    \item[] Question: Does the paper describe potential risks incurred by study participants, whether such risks were disclosed to the subjects, and whether Institutional Review Board (IRB) approvals (or an equivalent approval/review based on the requirements of your country or institution) were obtained?
    \item[] Answer: \answerNA{}
    \item[] Justification: The research does not involve human subjects or any procedure requiring IRB approval.

\item {\bf Declaration of LLM usage}
    \item[] Question: Does the paper describe the usage of LLMs if it is an important, original, or non-standard component of the core methods in this research? Note that if the LLM is used only for writing, editing, or formatting purposes and does \emph{not} impact the core methodology, scientific rigor, or originality of the research, declaration is not required.
    \item[] Answer: \answerNA{}
    \item[] Justification: Large language models are not a component of the core methodology; any use is limited to writing assistance, which does not require declaration per the NeurIPS LLM policy.

\end{enumerate}

\end{document}